\documentclass{article}

\usepackage[main, preprint]{mystyle}

\usepackage[utf8]{inputenc} % allow utf-8 input
\usepackage[T1]{fontenc}    % use 8-bit T1 fonts
\usepackage{natbib}
\usepackage{url}            % simple URL typesetting
\usepackage{booktabs}       % professional-quality tables
\usepackage{amsfonts}       % blackboard math symbols
\usepackage{nicefrac}       % compact symbols for 1/2, etc.
\usepackage{microtype}      % microtypography
\usepackage[table]{xcolor}         % colors
\usepackage{subcaption}
\usepackage{setspace}
\usepackage{graphicx}
\usepackage{multirow}
\usepackage{amsmath}
\usepackage{adjustbox}
\usepackage{caption}
\usepackage{enumitem}
\usepackage{makecell}
\usepackage{colortbl}
\usepackage{wrapfig}
\usepackage{pifont}
\usepackage[most]{tcolorbox}
\newtcolorbox{leftquote}{blanker, borderline west={2pt}{0pt}{black!60}, left=12pt, top=4pt, bottom=4pt}
\usepackage{algorithm}          % For algorithm float environment
\usepackage{listings}           % For code-style pseudocode
\usepackage{marvosym}
\usepackage{bxcoloremoji}
\usepackage{subcaption}
\setlist[itemize]{leftmargin=*}
\definecolor{mydarkblue}{rgb}{0,0.08,0.45}
\usepackage{amsmath,amsfonts,bm}

\def\eqref#1{equation~\ref{#1}}
\def\1{\bm{1}}

\DeclareMathAlphabet{\mathsfit}{\encodingdefault}{\sfdefault}{m}{sl}
\SetMathAlphabet{\mathsfit}{bold}{\encodingdefault}{\sfdefault}{bx}{n}

\usepackage{amssymb}
\usepackage{stmaryrd}

\usepackage{colortbl}
\definecolor{gaincell}{RGB}{232,244,236}
\definecolor{weakgaincell}{RGB}{241,248,243}
\definecolor{losscell}{RGB}{250,235,235}
\definecolor{basecell}{RGB}{241,242,244}
\definecolor{mygray}{HTML}{808080}
\definecolor{mydarkgreen}{HTML}{1B7F3A}

\usepackage{tabularx}
\usepackage{titlesec}
\titlespacing{\paragraph}{0pt}{0.4em}{0.5em}
\usepackage{hyperref}
\usepackage{mathtools}
\usepackage{amsthm}
\usepackage{algpseudocode} % algorithmicx

\usepackage{array}
\usepackage{caption}
\newcolumntype{g}{>{\columncolor{gray!10}}c}
\definecolor{step1color}{HTML}{D5E8D4} % Light Green
\definecolor{step2color}{HTML}{A8D08D} % Medium Green
\definecolor{step3color}{HTML}{82B366} % Darker Green
\definecolor{step4color}{HTML}{5A8A42} % Even Darker Green
\definecolor{promptcolor}{HTML}{F5F5F5} % Light Gray for prompt background
\definecolor{cond}{HTML}{2E75B6}
\definecolor{hdpocolor}{HTML}{2E75B6}
\definecolor{metiscolor}{HTML}{F3B000}
\definecolor{gainnum}{RGB}{55,125,88}
\definecolor{lossnum}{RGB}{176,76,82}
\definecolor{basenum}{RGB}{125,128,136}
\usepackage[dvipsnames]{xcolor}
\usepackage[normalem]{ulem}

\title{Business Arena: Benchmarking LLM Agents in a Realistic Marketplace}

\author{
    \textbf{Yijun Pan}$^{1,2\dagger}$ \quad
    \textbf{Yukun Lian}$^{1}$ \quad 
    \textbf{Kunyu Shi}$^{1}$ \quad 
    \textbf{Junbo Li}$^{1}$ \quad
    \textbf{Hongwei Xue}$^{1}$ \\
    \textbf{Sicong Xie}$^{1}$ \quad  
    \textbf{Guannan Zhang}$^{1}$ \quad 
    \textbf{Xiaoying Xing}$^{1\ddagger}$ \\
    $^1$Accio Team, Alibaba Group~~
    $^2$Yale University
}

\headerlogos{
    \includegraphics[height=0.5cm]{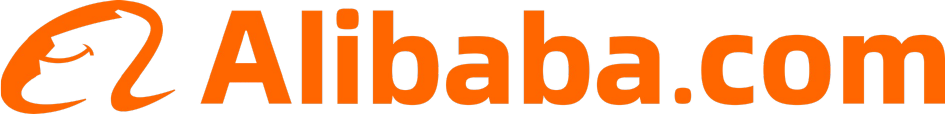}\hspace{0.3cm}
    \includegraphics[height=0.5cm]{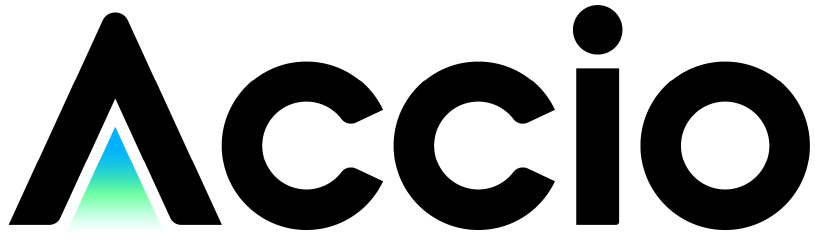}
}

\begin{document}

\maketitle

% [2] = †
% [3] = ‡
\footnotetext[2]{Work done during internship at Accio.}
\footnotetext[3]{Corresponding Author.}

\begin{abstract}
Running a business is a challenging form of intelligent work. Operators must infer opportunities from partial signals, commit capital under uncertainty, adapt to delayed outcomes in a changing market, and satisfy regulatory obligations before trading legally. Frontier LLM agents can increasingly complete complex workflows, suggesting their potential for end-to-end business operation, yet business-related capabilities are rarely evaluated in existing agent benchmarks. We introduce \textbf{Business Arena}\footnote{Project page: \url{https://business-arena.site.accio.ai}}, a controlled environment where an AI agent runs a cross-border shop, buying from suppliers and selling to buyers over a long horizon. We ground the arena in real Alibaba.com sourcing data and market conditions calibrated from authoritative sources. 
Delayed and coupled consequences make individual business decisions difficult to judge, but their combined outcome is measurable through final profit. Because profit alone cannot explain why an agent succeeds or fails, we compare agents with human-designed strategies to estimate available opportunity, use skill-level metrics to reveal underlying strengths and weaknesses, and trace realized gains and losses to the actions that produced them. Finally, we use mechanism ablations to establish that strong arena results reflect genuine business intelligence rather than neglect or simulator-specific shortcuts.
We evaluate 15 frontier models and find a ninefold difference in mean final net worth. Even the best-performing model falls substantially behind human-designed strategies, indicating that operating a business remains challenging for existing LLM agents. Skill-level analysis reveals recognizable operating styles, from margin-focused premium sellers to high-turnover wholesalers and customer-service specialists, while action-level attribution identifies the concrete sourcing, pricing, and recovery decisions that create or destroy value. Together, Business Arena takes a first step toward a realistic and trustworthy testbed for evaluating end-to-end business agents.
\end{abstract}

\begin{figure*}[h!]
    \centering
    \includegraphics[width=0.8 \textwidth]{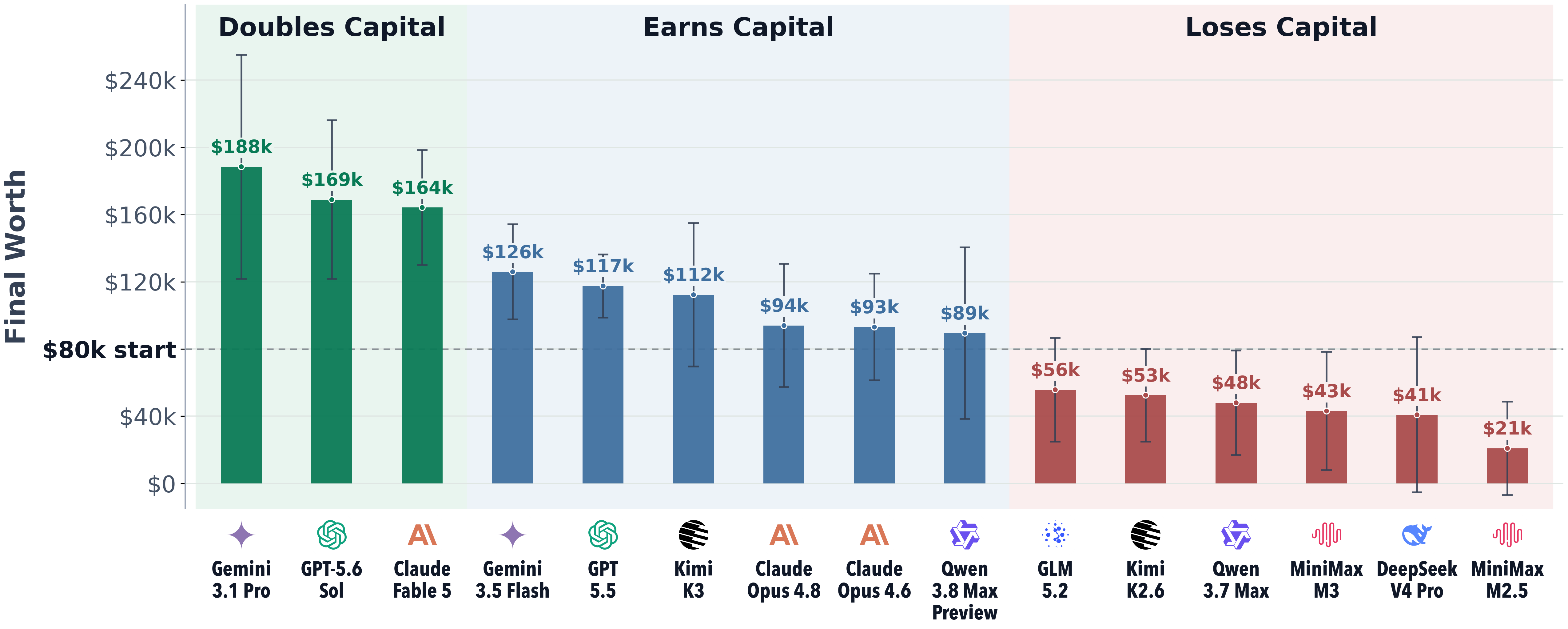}
    \caption{\textbf{Model performance in Business Arena.}
    Over the same long horizon, the strongest models more than double their capital, a middle group earns modest returns, and the weakest ones finish with less than they started.}
    \label{fig:main-model-performance}
\end{figure*}

\section{Introduction}
\label{sec:intro}

Recent advances in LLM agents are fueling a rapidly expanding industry. Recent research reports that average reasoning-token consumption per organization increased by approximately \(320\) times over the preceding year.\footnote{OpenAI, ``The State of Enterprise AI 2025.'' \url{https://openai.com/business/guides-and-resources/the-state-of-enterprise-ai-2025-report/}} Agents can write code, conduct research, operate software, and complete increasingly complex workflows~\cite{jimenez2024swebench,zhou2024webarena,yao2025taubench}. Together, these capabilities can support many individual business functions. Integrating them raises a more consequential prospect: an agent that operates the business itself, coordinating decisions across functions and directly determining how capital is deployed and profit is earned. This prospect motivates evaluating agents as end-to-end economic operators rather than only as tools for completing individual workflows.

What prevents such agents from operating real businesses today? Unlike other workflows, business requires high-stakes decisions under four major challenges. \textbf{Evidence is noisy}: true demand, competitor intentions, and customer preferences are hidden behind partial and sometimes conflicting market signals. \textbf{Feedback is delayed and difficult to attribute}: agents must commit capital before sales occur, while later outcomes rarely reveal which earlier decision was responsible. \textbf{The environment changes}: demand, costs, and competitor behavior continue to evolve, so previously sound plans can quickly become outdated. \textbf{Operational obligations persist}: businesses must satisfy compliance requirements, serve customers reliably, and cover recurring operating costs; neglecting any of these can erase the gains from otherwise sound commercial decisions. An agent must therefore do more than execute a fixed sequence of actions; it must continually interpret the market, commit resources, revise its operation, and satisfy obligations as new evidence arrives. 

These challenges make direct deployment in a live business an unsuitable first test. Model errors can waste real capital, mislead customers, or violate regulations; moreover, losses cannot be undone, and market conditions cannot be held constant across runs for reproducible evaluation. A credible first evaluation must preserve the challenges of real business operation without placing real firms at risk. Existing agent benchmarks provide limited coverage of business scenarios~\cite{jimenez2024swebench,zhou2024webarena,yao2025taubench,desai2026swemarathon}, while recent business-oriented benchmarks either capture only part of the business loop or simplify the challenging conditions that make business difficult~\cite{backlund2025vendingbench,wang2024shopbench,he2026ycbench,chen2026ceobench}.

We therefore introduce \textbf{Business Arena}, a controlled environment in which an AI agent independently operates a cross-border business-to-business shop over a long horizon. This setting exposes an end-to-end business loop spanning market research, sourcing, inventory management, pricing, sales, customer service, compliance, and finance. We ground this loop in real data: products, supplier offers, prices, minimum order quantities, and lead times come from real Alibaba.com listings, while demand cycles, tariffs, and related market conditions are calibrated from authoritative sources. More than 60 tools span the seller loop: typed MCP calls support interleaved reasoning and action, while back-end APIs and a persistent workspace support model-authored analyses, scripts, and operating routines.

The arena's mechanisms preserve the challenges that make this loop difficult. The agent must infer product--market opportunities from partial and sometimes conflicting evidence, commit capital before results are known, and adapt as buyers, supplier costs, competitors and trade conditions evolve. At the same time, business activity is constrained by obligations that cannot be deferred: market entry requires approvals, buyers expect timely and accurate service, and operating costs accrue regardless of revenue. Business Arena therefore evaluates whether an agent can sustain a coherent business in a realistic and challenging environment.
 
Our evaluation is designed to ensure that arena scores reflect the business capabilities we seek to measure and to explain the behavior behind them. A realistic world alone does not guarantee this: agents might achieve high scores through simulator-specific shortcuts rather than competent business decisions. We therefore use mechanism ablations to compare intended behavior with neglect and misuse, testing whether stronger arena performance corresponds to stronger business intelligence. Even a meaningful final score remains too coarse to explain why an agent succeeded or failed. We decompose performance into skill-level metrics that expose capabilities and operating patterns hidden by aggregate outcomes, and implement an attribution toolkit that traces realized gains and losses back to the actions that produced them. This provides fine-grained credit assignment for model analysis and future training-data construction.

We evaluate 15 frontier models and find that mean final net worth ranges from \$20{,}856 to \$188{,}488, a \(9.0\) times gap. The arena remains challenging: \(51\%\) of all runs lose money, and only four models preserve their starting capital in every trial. The strongest expert-designed strategy earns more than twice the best model mean, revealing substantial headroom. Successful models combine disciplined capital deployment, sell-through, margin-preserving pricing, and continued market learning, whereas weaker models leave capital idle, destroy margin, or incur compliance violations. Models with similar final outcomes can nevertheless differ in how they interpret market evidence and how reliably they translate those beliefs into coordinated actions. Some protect margins through selective pricing, while others prioritize sell-through and capital turnover, reflecting strategies recognizable among real sellers. By revealing both how models understand the market and how their decisions create or destroy value, Business Arena provides a controlled testbed for evaluating end-to-end business agents.

\section{Related Work}

\subsection{Long-Horizon Agent Evaluation}
\label{sec:related-longhorizon}

LLM agent evaluation has progressed from short-horizon, verifiable tasks (code patching~\cite{jimenez2024swebench}, web navigation~\cite{zhou2024webarena}, policy compliance~\cite{yao2025taubench}) to long-horizon benchmarks such as SWE-Marathon~\cite{desai2026swemarathon}, which requires agents to navigate entire repositories and iterate through failing test suites over long testing horizons. These longer benchmarks confirm that frontier models can sustain coherent execution, but their environments remain fundamentally \emph{fixed}: success is binary, a correct answer exists, and the world does not change while the agent acts. The next evaluation frontier demands environments that are noisy, non-stationary, and lack a single verifiable solution; where feedback is delayed, outcomes couple across decisions, and the world evolves whether the agent acts or not. Among such domains, business is one of the most economically consequential and challenging.

\subsection{Business Simulation}
\label{sec:related-bizsim}

Business simulations have long tested multidimensional decision-making under uncertainty~\cite{keys1990management,sterman1989modeling}. Their central tension is between realism and evaluability: real markets contain hidden state, changing conditions, coupled decisions, and no single ground-truth trajectory, while simplified environments are easier to evaluate but remove much of what makes business difficult. Agent-based computational economics further shows that heterogeneous actors can produce market dynamics absent from equilibrium models~\cite{tesfatsion2006agent}.

Recent benchmarks capture only parts of this problem. VendingBench~\cite{backlund2025vendingbench}, ShopBench~\cite{wang2024shopbench}, and YC-Bench~\cite{he2026ycbench} study narrower commercial workflows, while CEO-Bench~\cite{chen2026ceobench} focuses on managing a simulated SaaS company. None jointly evaluates the full physical-commerce cycle, where sourcing, inventory, landed costs, pricing, demand, customer service, logistics, and compliance interact. Business Arena addresses this gap through a data-grounded, autonomously operated marketplace paired with diagnostic evaluation of the opportunities available and the decisions that capture them.

\section{Designing a Marketplace for Business Intelligence}
\label{sec:arena-overview}

Business Arena places one agent-operated shop in a marketplace of suppliers, buyers, and competing sellers. The agent is responsible for operating the business end to end and adapting it over time. We design the arena's mechanisms and scenarios around the challenges of real business operation, allowing us to evaluate how well an agent manages the resulting business over time.

\subsection{Episode Lifecycle}
\label{sec:mkt-operation}

An episode begins with a pre-opening setup phase. The agent declares an initial shop focus, which it may revise later as market evidence accumulates, then inspects available opportunities and decides whether to acquire an initial inventory portfolio or retain capital for later sourcing. Inventory selected during setup is available on day 0, allowing the shop to begin trading without an unavoidable shipping delay. After setup, the agent clears the requirements of the markets it intends to enter and configures the listings through which its products are offered.

During each simulated day, the agent may inspect the market, adjust its offers, source or replenish inventory, allocate advertising spend, manage liquidity, and respond to buyers and suppliers before choosing to advance the day. The arena then advances the rest of the market autonomously: buyers make purchasing decisions, competitors update their operations, shipments progress, financial costs are deducted, and new events or market signals may emerge. Because these processes continue independently of the agent, the conditions it faces change throughout the episode. The day's outcomes update the persistent world state, which the agent can inspect before deciding whether to scale successful products, revise prices, change focus, or liquidate weak inventory. After the final day, remaining inventory contributes to final net worth at a discounted salvage value.

\subsection{Arena Mechanisms}

\begin{figure*}[h]
    \centering
    \includegraphics[width=\textwidth]{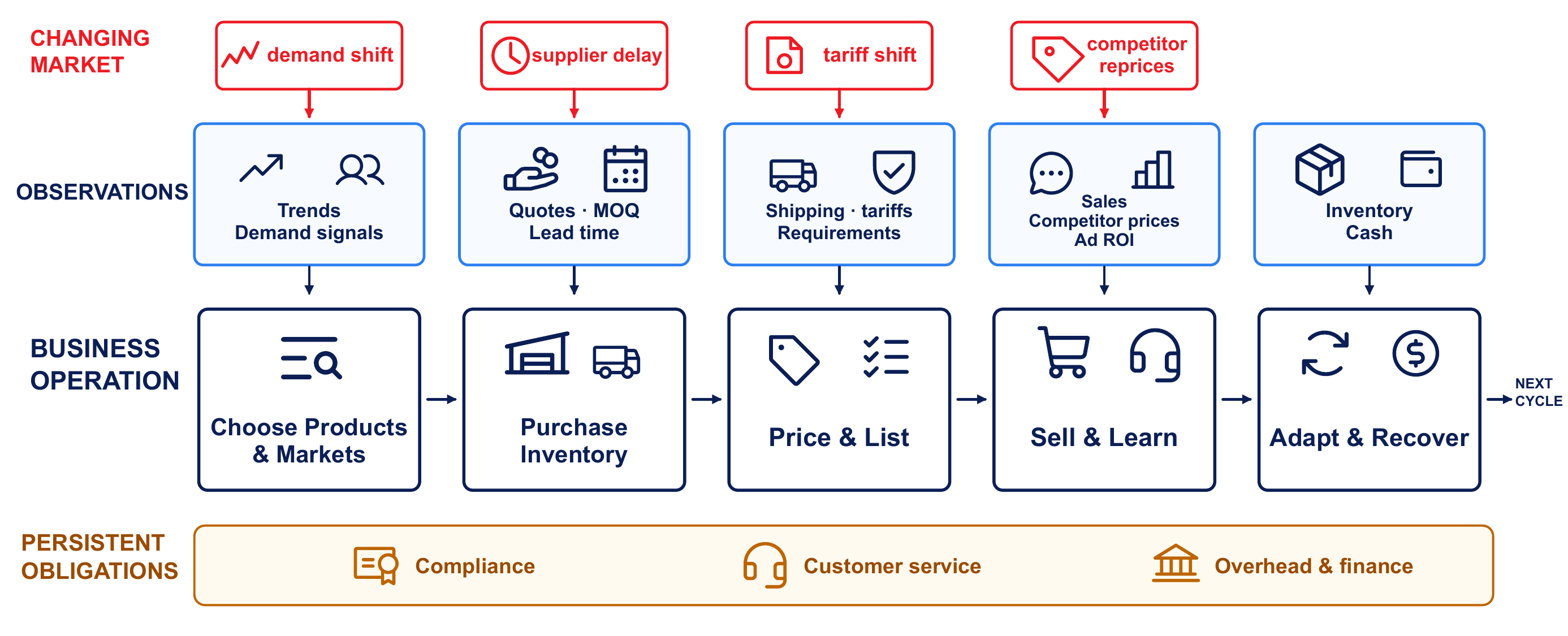}
    \caption{Overview of arena design. The agent selects markets, purchases inventory, prices and lists products, learns from sales, and adapts its operation. It acts on \textbf{partial observations} while supplier disruptions, competitor repricing, and demand shifts create a \textbf{changing market}. Meanwhile, \textbf{operational obligations} persist and economic feedback remains \textbf{delayed}.}
    \label{fig:business-agent-loop}
\end{figure*}

Two principles guide the arena's design. First, its mechanisms are built around the central challenges of real business operation. Second, it covers the major components of an end-to-end business cycle, allowing decisions in one part of the business to affect outcomes elsewhere. Together, these choices aim to preserve the key abstractions of a challenging business world within a controlled environment. We show an overview of the arena in figure \ref{fig:business-agent-loop}.

\paragraph{Incomplete and noisy evidence.}
Real markets rarely reveal a ground-truth state: evidence is scattered and noisy, and its relationship to future outcomes is uncertain. Business Arena preserves this uncertainty without reducing the task to guesswork. The agent cannot directly observe latent market conditions, but it can gather evidence from public signals, competitors, counterparties, and its own operating history. The way arena exposes evidence about demand illustrates this design. Festival calendars provide relatively explicit timing and signals, testing whether agents retrieve and act on available information. Google Trends data require agents to compare lagged historical patterns and infer whether interest is rising or falling. Market events are harder: precursors may be real or false, but sufficient public evidence is available for careful agents to distinguish stronger signals from rumors. Across these levels, the information needed for informed decisions exists, but the agent must decide what to inspect, what to trust, and when the evidence is strong enough to act.

\paragraph{Delayed economic consequences.}
The agent must decide what kind of shop to build, which opportunities to pursue, how much capital to deploy, and how broadly to diversify, yet the consequences of these choices emerge only after a delay. Several mechanisms make this planning problem consequential: fixed daily overhead makes leaving capital idle costly, while inventory holding fees and final salvage discounts penalize reckless deployment into products that do not sell. Shipping, tariffs, and platform commissions further determine whether an apparently attractive sale remains profitable after the full cost stack. Business Arena also preserves the flexibility to recover from weak decisions: agents can change their shop focus or liquidate inventory and redeploy the recovered capital, but both actions come at a cost.

\paragraph{A changing market.}
The marketplace continues to evolve whether or not the agent acts. Buyers choose among all available sellers, while competitors independently update their prices, inventory and advertising budgets. Demand, supplier costs, events, and trade conditions also change during the episode. We also introduce realistic market events, including shifts in demand and supplier costs, trade-policy shocks calibrated from real U.S.--China tariff changes, shipping disruptions, and new compliance requirements. A product, price, or market that was attractive earlier may therefore become less viable as the episode unfolds.

\paragraph{Persistent operational obligations.}
Identifying profitable opportunities is not enough for real business: it must satisfy regulatory and operational obligations before it can trade legally and fulfill orders reliably. Business Arena introduces market-specific compliance requirements: agents must identify the necessary approvals, apply early enough for them to clear, and delay entry until they are granted; otherwise, illegal trading incurs substantial fines. Buyers also arrive with offers and factual questions that require timely and accurate responses. Recurring operating costs continue regardless of whether the agent is actively expanding. These mechanisms ensure that commercial gains can be reduced or erased when the agent neglects compliance, customer service, or basic operating discipline.

\paragraph{End-to-end business coverage.}
The arena covers the major stages of the business cycle: market research and opportunity selection; sourcing, inventory, and logistics; market entry and compliance; pricing, advertising, and sales; customer and supplier interactions; and financial management and recovery. Recent business benchmarks capture important parts of this process but generally focus on narrower operating surfaces~\cite{backlund2025vendingbench,wang2024shopbench,he2026ycbench,chen2026ceobench}. By bringing these components into one marketplace, Business Arena more closely reflects the scope of end-to-end business operation that an autonomous seller would face in practice.

\subsection{Business Capabilities Evaluated}

The arena mechanisms create the challenges of the business world, while in this section we identify the capabilities agents need to navigate this world. Prior business and management literature characterizes business operation as making decisions without complete knowledge~\cite{simon1955behavioral} while repeatedly identifying opportunities, committing resources, and adapting operations as conditions evolve~\cite{teece1997dynamic}. Based on these accounts, we organize business intelligence into the following four capabilities. Appendix~\ref{sec:orch-tools} provides a detailed mapping between each arena mechanism to the business capabilities it evaluates and the tools available to the agent.

\paragraph{Decision-Making Under Uncertainty.}
Before a seller can decide what to buy or where to sell, it must form a view of what is happening in the market. A strong strategy actively gathers and cross-checks available signals, tests its expectations against realized outcomes, and updates its beliefs when the evidence changes. A weak strategy follows the latest signal uncritically, trusts claims without verification, or continues acting on a stale belief despite contrary sales and inventory outcomes.

\paragraph{Strategic Planning Under Constraints.}
After forming a view of the market, the seller must turn it into an operating plan: what kind of shop to build, which opportunities to pursue, how much capital to deploy, and how broadly to diversify. A strong strategy coordinates product selection, market entry, capital allocation, and risk while preserving enough flexibility to respond to later evidence. A weak strategy leaves capital idle, concentrates on unsupported opportunities, accumulates excessive inventory, or continues following its initial plan after the market has changed.

\paragraph{Insight-to-Action Alignment.}
A sound business plan is not enough; the seller must get the operational details right. It must translate its plan into concrete choices about suppliers, order quantities, prices, market entry, and timing while accounting for current costs and requirements. A strong strategy checks the relevant information, performs the necessary calculations, and revises its actions when the economics change. A weak strategy understands the direction of a decision but uses stale inputs, omits important costs, or fails to carry the plan through correctly.

\paragraph{Cooperation \& Competition.}
A seller does not operate in isolation. Buyers, suppliers, and competing sellers pursue their own interests, and the agent must work with them without losing sight of its own objective. A strong strategy monitors competitive moves, learns what buyers value, knows when to accept, counter, or walk away, and provides accurate service throughout the transaction. A weak strategy discounts mechanically, relies on templated or unsupported responses, or continues operating as though other participants were static.

\subsection{An Interface for Diverse Strategies}

Business Arena provides a flexible operating interface for agents to carry out realistic and diverse strategies. More than 60 tools span the business cycle, but they are not limited to narrow, predetermined actions. Tools expose flexible arguments that let agents construct their own workflows. For example, rather than returning an impractically large supplier list, supplier search allows agents to define filters and rankings over price, quality, MOQ, lead time, and other attributes. The action space also captures multiple ways of implementing the same business decision: an agent may directly update a listing price or define quantity-based price tiers, after which the market matches each buyer against the applicable offer. We also provide flexible tool orchestration: typed MCP calls support interleaved reasoning and individual actions, while back-end APIs and a persistent workspace enable larger analyses, scripts, and autonomous routines. This interface exercises foundational agent capabilities, including tool-use proficiency, cross-source evidence gathering, long-horizon state management, and reliable multi-step execution. Rather than evaluating these abilities in isolation, Business Arena tests whether agents can combine them to sustain an evolving business.

Each agent runs as an unprivileged user in an isolated OpenClaw sandbox. It may freely read, write, and execute programs in its own workspace, while the arena service runs under a separate identity whose source code, database, and hidden state are protected through filesystem permissions. Agents can therefore organize and automate their business freely while interacting with the market only through its public interfaces.

\section{Diagnosing Model Performance}
\label{sec:evaluation}

Business performance is the product of market beliefs, capitol commitments, quantitative decisions, and interactions with other market participants. This complexity is what the arena is intended to test, but it also makes a final score difficult to interpret: similar outcomes can arise from different operating choices, and the value of those choices depends on the opportunities present in the market. We therefore bound evaluation around two questions: \textbf{what opportunity was available in this marketplace?} and \textbf{how much of that opportunity did the agent capture, through which decisions?} Terminal net worth remains the leaderboard outcome, but serves as the starting point for analysis.

\subsection{Estimating Available Opportunity}
\label{sec:evaluation-opportunity}

Because an open-ended market has no single correct trajectory, a model's final score needs a credible reference for what could have been achieved under the same conditions. We construct a library of deterministic strategies using only information available to the evaluated agents. These are not isolated feature heuristics or locally optimal rules stitched together. They represent coherent operating strategies in which decisions across the business reinforce one another. For example, our leading expert-designed strategy maintains Bayesian estimates of demand and route contribution, repeatedly updates them from public evidence, sales, and inventory exposure, and uses the resulting estimates to coordinate pricing, advertising, replenishment, and capital redeployment. These strategies therefore represents a broad range of plausible seller behavior, providing both an empirical estimate of available opportunity and concrete operating traces against which model decisions can be compared. Appendix~\ref{app:strategy-reserve} describes its architecture and coverage.

\subsection{Attributing Captured Value}
\label{sec:evaluation-attribution}

A final business outcome compresses many capabilities and decisions into one number. We therefore analyze performance at two resolutions: skill-level metrics reveal where an agent is strong or weak, while action-level attribution links realized gains and losses to concrete decisions. Stateful evaluation complements these diagnostics by restoring the exact context around a decision, allowing alternative continuations to be compared from the same business state.

\paragraph{Skill-level diagnosis.}
For each business capability and operational subtask, we define economically grounded submetrics that summarize how well the agent performed. For example, advertising is evaluated through full-funnel return on advertising spend (ROAS) and advertising return on investment (ROI), which measure whether the agent directs spending toward listings that convert paid exposure into profitable downstream orders. Similar submetrics characterize opportunity selection, capital deployment, customer service, compliance and so on. The resulting profiles reveal strengths, weaknesses, and recognizable operating styles in reality that would remain hidden behind a single final score.

\paragraph{Action-level attribution.}
Skill-level metrics identify where an agent performed well or poorly, but improving that capability requires a finer signal about which decisions should be reinforced or corrected. We therefore implement attribution tools that trace every realized gain and loss through recorded economic transitions to the model actions that produced it. Each sourcing, pricing, service, financing, or recovery action can consequently be paired with its downstream economic contribution. These action-linked outcomes provide dense credit assignment for model diagnosis and can support future training-data curation and reinforcement learning.

\paragraph{Stateful evaluation.}
Action-level attribution identifies decisions associated with gains and losses, but improving a model also requires testing how outcomes change when those decisions are replaced. Such comparisons must begin from the same business state; otherwise, outcome differences may reflect different preceding histories. Business Arena therefore implements a save--fork--load pipeline that jointly restores the model-visible context, OS-level workspace, and exact marketplace state. Evaluators can continue alternative models, actions, or reference policies from the same checkpoint, enabling like-for-like comparisons, reusable difficult-state tests, and test-time scaling through selective continuation. Appendix~\ref{app:stateful-evaluation} describes the implementation and applications.

\section{Experiments}
\label{sec:experiments}

\paragraph{Testing Environment.}

Each evaluation runs in a fresh, isolated sandbox with a preinstalled OpenClaw runtime. The model receives agent-facing documentation as a layered skill set: a compact top-level guide introduces the task and core tool surface, while deeper domain files provide detailed market and operating information when needed. Filesystem permissions prevent the model from accessing the arena source code, database, or hidden simulator state.

\paragraph{Models Evaluated.}
\label{sec:exp-leaderboard}

We evaluate frontier models spanning both proprietary and open-weight systems. The proprietary cohort includes GPT 5.6 Sol, GPT-5.5, Claude Fable 5, Opus 4.6 and 4.8, Gemini 3.1 Pro and 3.5 Flash, and Qwen 3.7 Max. The open-weight cohort includes GLM 5.2, Kimi K2.6 and K3, DeepSeek V4 Pro, MiniMax M2.5 and M3, and Qwen-3.8-Max-Preview. Leaderboard results are mean final net-worth across \textbf{10} runs. All models receive maximum thinking effort, specifically GPT 5.6 Sol is set to pro reasoning mode.

\section{Results}
\label{sec:results}

\subsection{Performance Landscape}
\label{sec:results-leaderboard}

\begin{figure}[t]
    \centering
    \includegraphics[width=\linewidth]{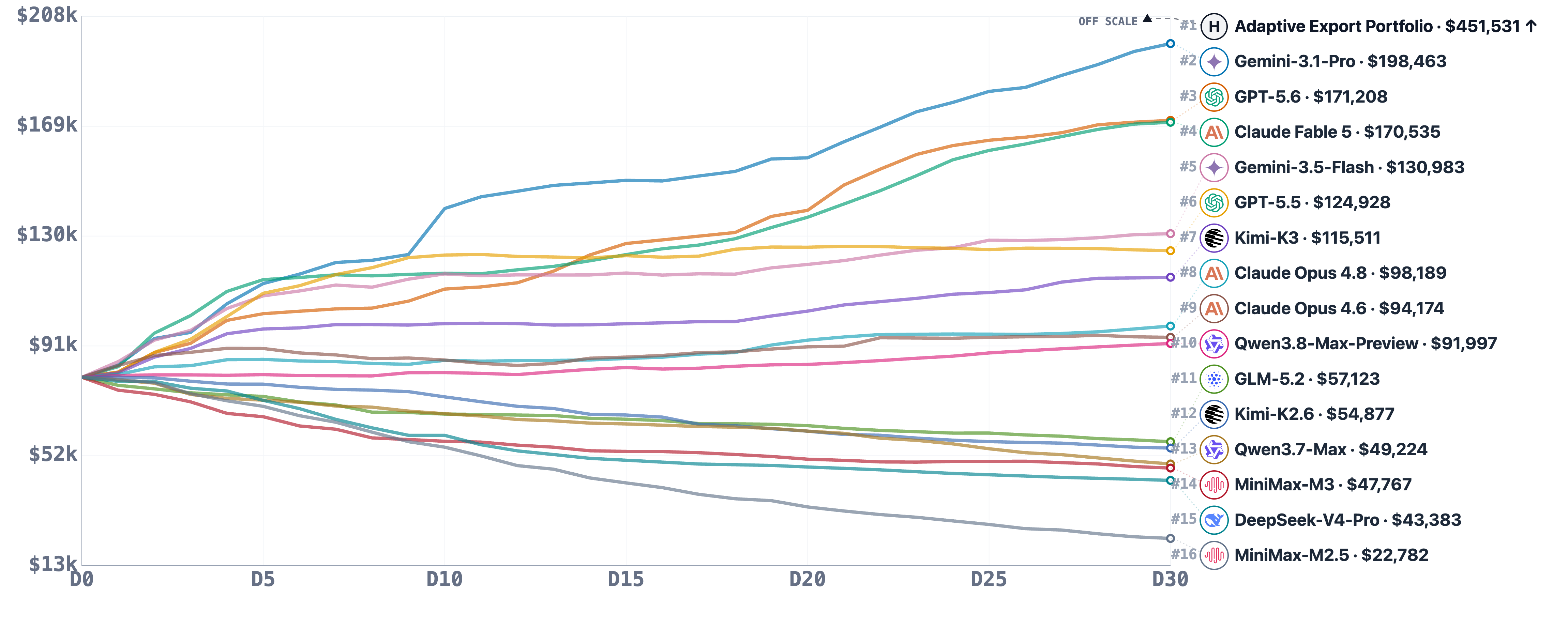}
    \caption{Main leaderboard over 15 model families, averaged across ten runs under the same world condition. Dashed lines denote expert-designed strategies that use only agent-visible information.
    }
    \label{fig:leaderboard}
\end{figure}

Frontier models differ sharply on Business Arena. Across 15 models, mean final net worth ranges from \$188{,}488 for Gemini 3.1 Pro to \$20{,}856 for MiniMax M2.5, a \(9.0\times\) difference, and \(51\%\) of runs lose money relative to the \$80{,}000 starting capital. Only four models preserve their starting capital in every trial, showing that profitable operation remains inconsistent even among frontier agents. This matters in practice because a commercially useful agent must preserve capital consistently rather than offset frequent losses with a few exceptional runs. The reported ten-run means are reliable (\(\mathrm{ICC}=0.944\))~\citep{shrout1979intraclass}, while disjoint five-run subsets preserve the ranking (\(\rho=0.898\)) and recover the same leading group in \(98.7\%\) of cases. Business Arena therefore remains challenging while producing stable model comparisons (Appendix~\ref{app:variance} provides details of score variance).

Substantial headroom remains in Business Arena. The strongest expert-designed strategy reaches \$436{,}195 in the same world, more than twice the best model mean. It achieves this by coordinating market evidence with portfolio selection, sourcing, pricing, compliance, service, and capital allocation, then adapting as outcomes accumulate. Importantly, several other strategies resembling diverse real-world seller doctrines also perform well and surpass most models, indicating that the arena does not reward only a single narrow policy or simulator-specific shortcut (refer to appendix \ref{app:strategy-reserve} for details).

\subsection{How Models Run the Business}
\label{sec:results-profiles}

\begin{figure*}[h!]
    \centering
\includegraphics[width=0.8 \textwidth]{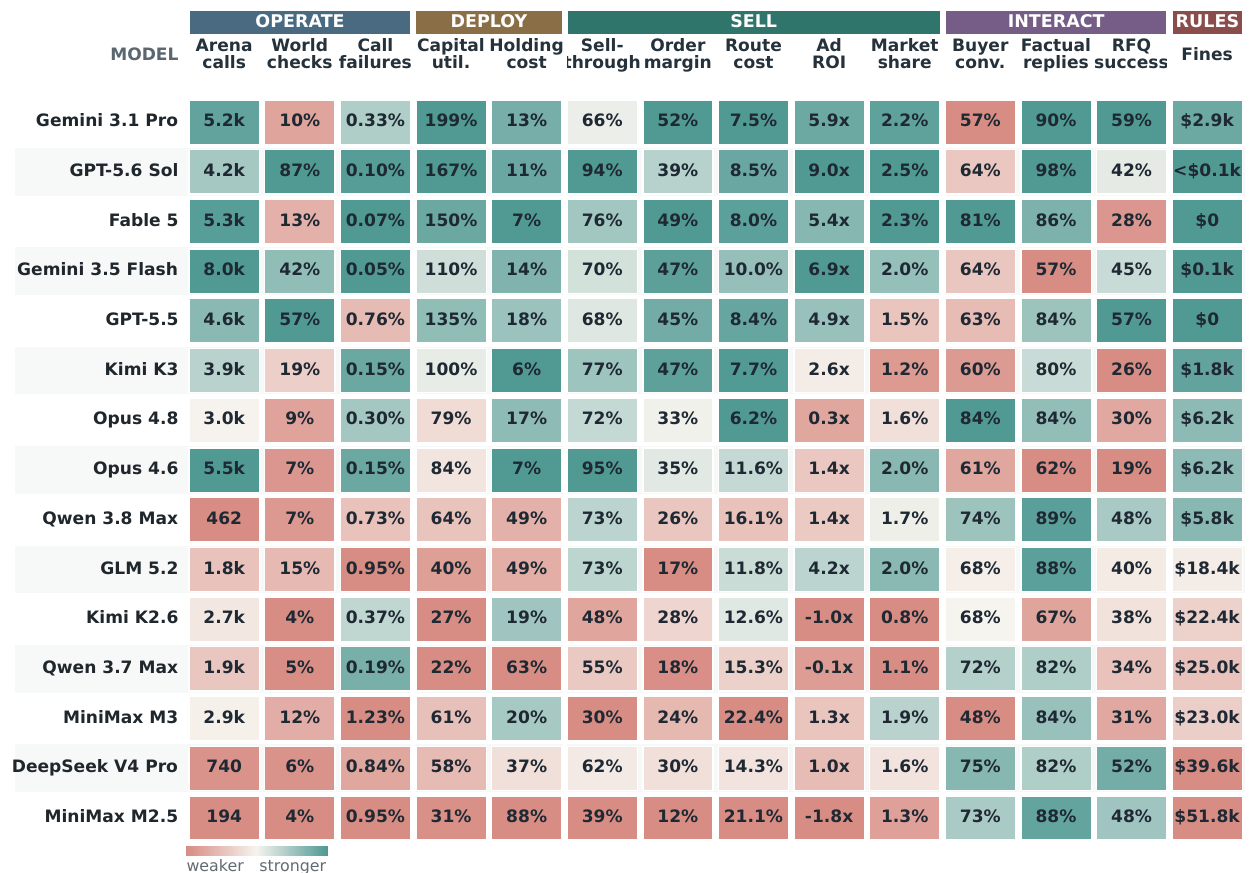}
    \caption{\textbf{Model diagnostic profiles.} Models exhibit different strengths across operating fluency, capital deployment, selling, customer interaction, and compliance. Colors indicate cohort-relative performance from weaker to stronger.}
    \label{fig:model-diagnostic-heatmap}
\end{figure*}

End-to-end business performance, both in practice and in Business Arena, results from many coupled decisions made over a long horizon. A single final net-worth score can therefore conceal important differences in model behavior. Figure~\ref{fig:model-diagnostic-heatmap} provides an overall decomposition of model performance into skill-level metrics (refer to Appendix \ref{app:model-diagnostics} for metric explanations). Leading models tend to combine active capital deployment, healthy selling economics, and reliable compliance, but none dominates every dimension. Lower-ranked models are generally more uneven, sometimes demonstrating strong customer interaction or tool reliability without converting those capabilities into profitable full-cycle operation. We next examine important skill-level results and use trajectory evidence to understand the model behaviors behind them.

\begin{figure*}[t]
    \centering
    \begin{subfigure}[t]{0.49\textwidth}
        \centering
        \includegraphics[width=\linewidth]{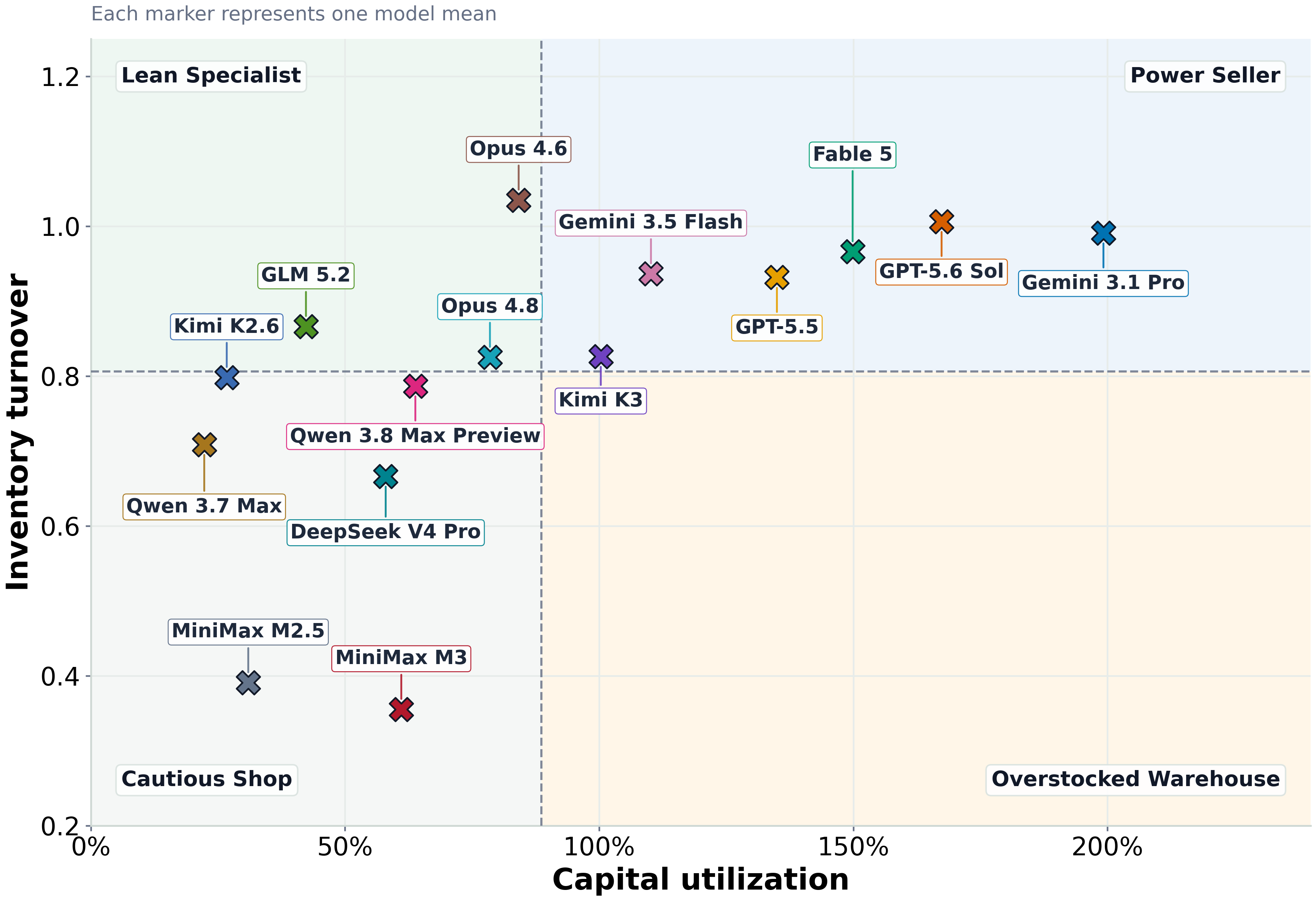}
        \caption{Capital deployment and inventory turnover.}
        \label{fig:capital-deployment}
    \end{subfigure}
    \hfill
    \begin{subfigure}[t]{0.49\textwidth}
        \centering
        \includegraphics[width=\linewidth]{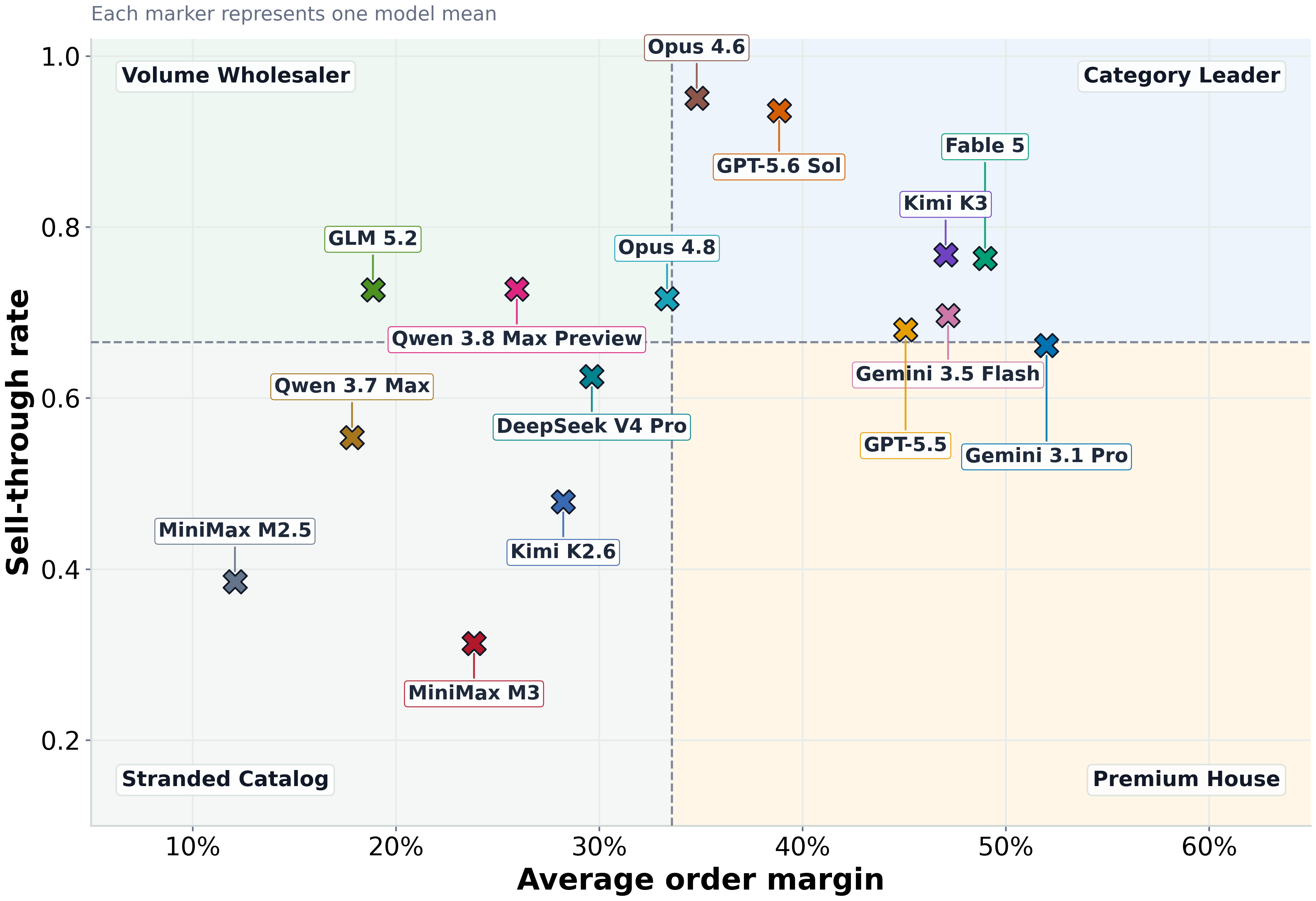}
        \caption{Margin and sell-through.}
        \label{fig:unit-economics}
    \end{subfigure}
    \caption{\textbf{Core operating trade-offs.} Successful agents identify opportunities and deploy capital into inventory that will sell (left), then preserve margin without pricing themselves out of the market (right). Crosses denote model-family means and lighter points individual runs.}
    \label{fig:business-skill-performance}
\end{figure*}

\paragraph{Capital utilization and turnover.}
Strong businesses find opportunities worth funding and repeatedly redeploy the capital recovered from sales. As shown in Figure~\ref{fig:capital-deployment}, Gemini 3.1 Pro, GPT-5.6 Sol, and Fable 5 reach cumulative capital utilization of 199\%, 167\%, and 150\%, respectively, while maintaining inventory turnover near 1.0. Weak models fail in two different ways. Qwen 3.7 Max resembles an \textbf{Underinvested Shop}, deploying only 22.3\% of starting capital and leaving much of the available opportunity unexplored. MiniMax M3 deploys more, but its 0.36 turnover leaves capital stranded in slow-moving inventory. Strong agents therefore do not simply spend more: they use realized sales to guide replenishment and recycle capital into opportunities supported by evidence.

\paragraph{Margin and sell-through.}
Once capital is deployed, agents must convert inventory into profitable sales. Figure~\ref{fig:unit-economics} reveals several viable operating styles. Gemini 3.1 Pro behaves like a \textbf{Premium House}, protecting a 52.0\% average order margin while accepting lower sell-through. GPT-5.6 Sol and Opus 4.6 resemble \textbf{Volume Wholesalers}, trading some margin for sell-through above 93\%. Fable 5 is the clearest \textbf{Well-Rounded Seller}, combining healthy margin and demand capture. MiniMax M2.5 instead resembles a \textbf{Stranded Shop}, earning only a 12.1\% average order margin while selling through less than 40\% of its inventory. In one trace, its pricing program assigns each SKU a single price across markets despite recognizing route-specific costs, causing 98 of 142 orders to fall below landed cost. The arena therefore permits different pricing strategies, but rewards models that preserve unit economics while keeping inventory attractive enough to sell.

\begin{figure*}[t]
    \centering
    \begin{subfigure}[t]{0.49\textwidth}
        \centering
        \includegraphics[width=\linewidth]{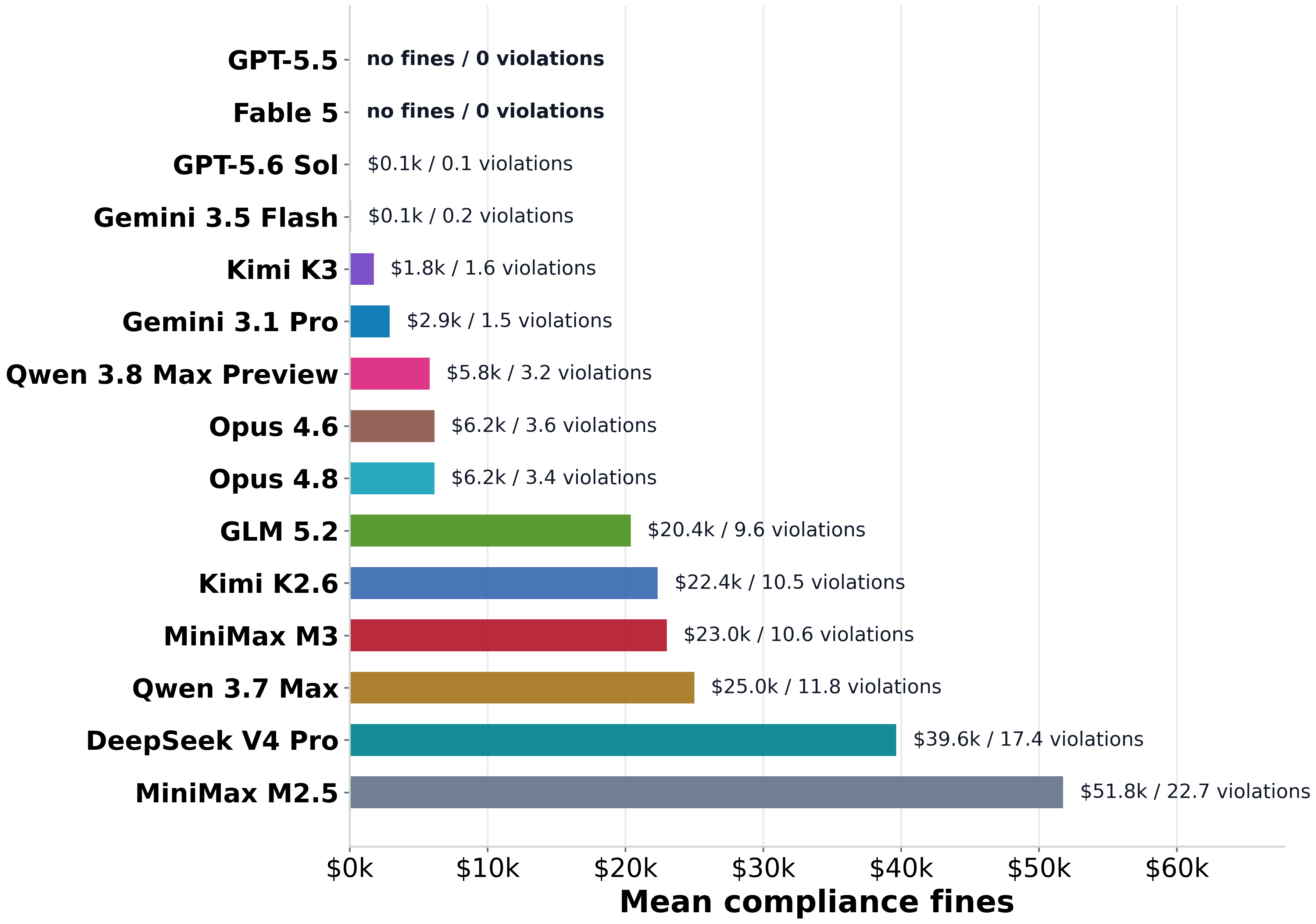}
        \caption{Compliance violations and resulting fines.}
        \label{fig:compliance-discipline}
    \end{subfigure}
    \hfill
    \begin{subfigure}[t]{0.49\textwidth}
        \centering
        \includegraphics[width=\linewidth]{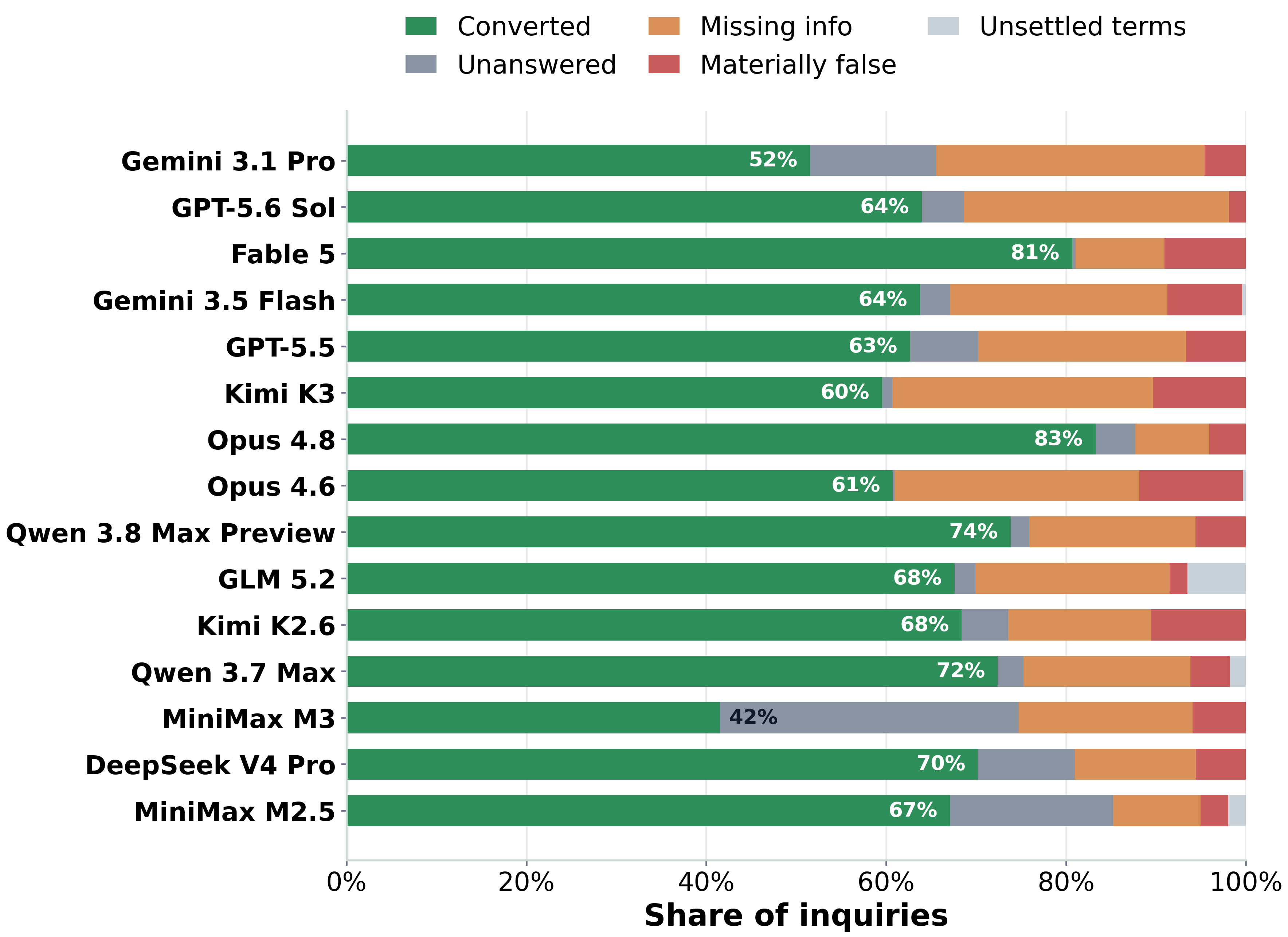}
        \caption{Customer-service outcomes.}
        \label{fig:customer-service}
    \end{subfigure}
    \caption{Compliance exposes deployment-critical reliability failures (left), while customer service reveals specialized strengths that do not follow the aggregate leaderboard (right).}
    \label{fig:full-cycle-skills}
\end{figure*}

\paragraph{Customer service.}
Customer-service results, shown in Figure~\ref{fig:customer-service}, reveal a different model ranking from final net worth. This task requires agents to infer buyer preferences while providing complete and factually supported product information. Opus 4.8 converts 84\% of inquiries, Fable 5 converts 81\%, and Qwen 3.8 converts 74\%. Opus 4.8 behaves more like a \textbf{Customer-Service Specialist} than a full-cycle business operator: despite its weak overall economics, it consistently checks relevant product and inventory information before responding to buyers. Conversely, Gemini 3.1 Pro converts only 57\% of inquiries and leaves substantially more unanswered or incomplete. These results expose customer-service strengths and weaknesses that are not visible in the overall leaderboard.

\paragraph{Compliance discipline.}
Business Arena treats compliance seriously: rather than merely recording violations, it imposes substantial fines when agents trade without required approvals. This reflects the financial and legal risks that an unreliable autonomous seller would create in practice. Shown in figure \ref{fig:compliance-discipline}, strong-performing models generally maintain better compliance: Fable 5 and GPT-5.5 incur no penalized violations, while GPT-5.6 Sol and Gemini 3.5 Flash are nearly clean. Weaker models often fail this requirement. MiniMax M2.5 averages \(22.7\) violations and \$51{,}750 in fines, while DeepSeek V4 Pro averages \(17.4\) violations and \$39{,}650. These failures show that profitability alone is insufficient for deployment: reliable compliance and safety behavior must be established before agents can be entrusted with real business operations.

\subsection{Model-authored workflows}
\label{sec:results-programs}

In this section, we examine model-authored business workflows to understand the intelligence behind the results. These examples show how models turn market evidence into sourcing and pricing plans and how they revise those policies when outcomes depart from their plans.

\begin{figure}[h!]
    \centering
    \includegraphics[width=\linewidth]{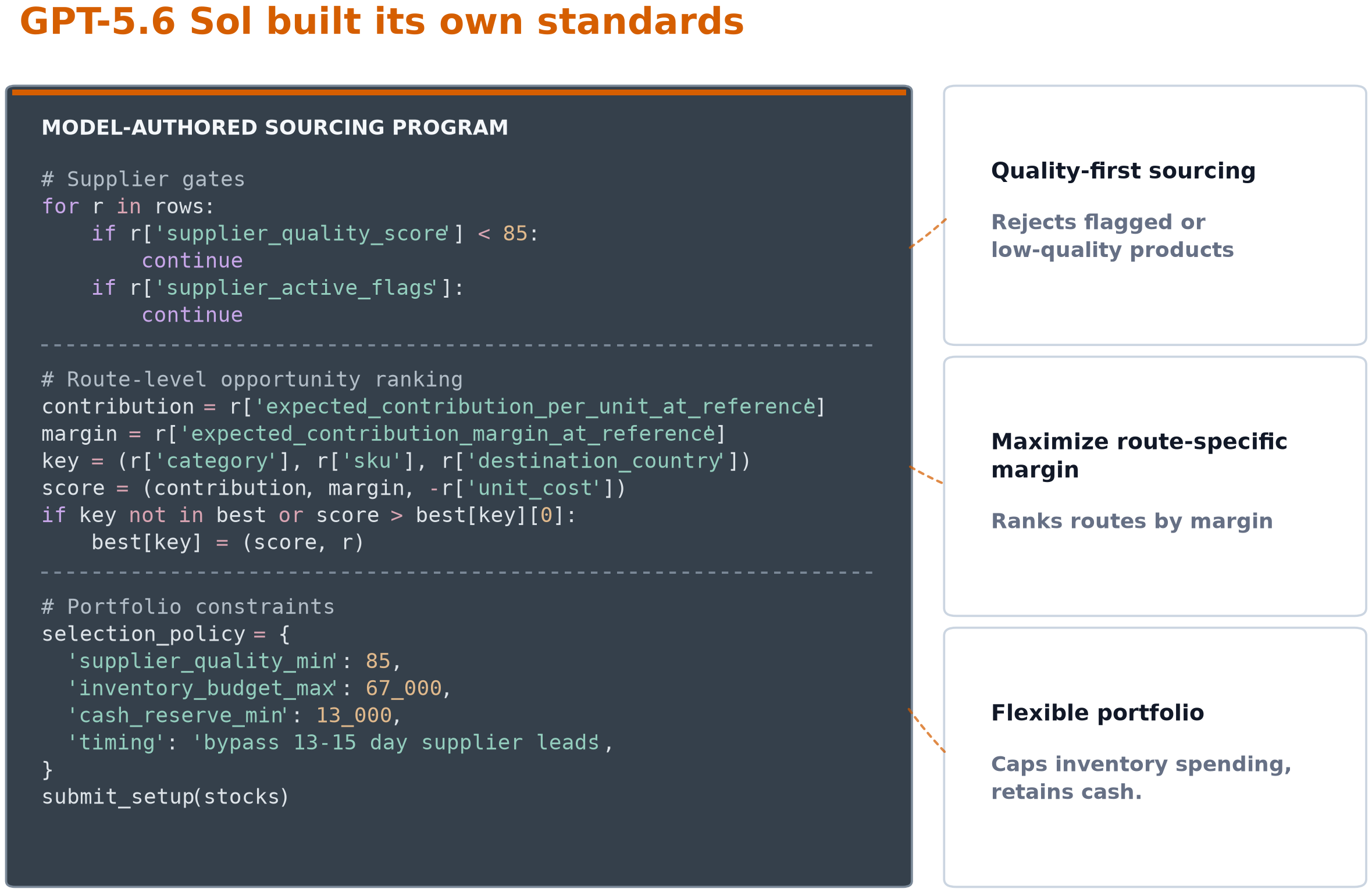}
\caption{\textbf{GPT-5.6 Sol behaves like a Volume Wholesaler.} Its self-defined sourcing standards reject risky products, prioritize high-margin routes, and diversify inventory while retaining cash, supporting aggressive capital deployment and high sell-through.}
    \label{fig:gpt56-supplier-policy}
\end{figure}

The preceding metrics characterize GPT-5.6 Sol as a \textbf{Volume Wholesaler}, combining high capital deployment with rapid inventory turnover. Figure~\ref{fig:gpt56-supplier-policy} reveals the sourcing policy behind this behavior. The model writes filters for supplier--category--country triplet, prioritizing combinations with high quality and demand, thus more likely to sell. After finding the opportunities, it built a diverse and aggressive spending plan: a nine-SKU portfolio spanning the 3 category it focuses on, and achieved high sell-through at the end.

\begin{figure}[h]
\centering
\includegraphics[width=\linewidth]{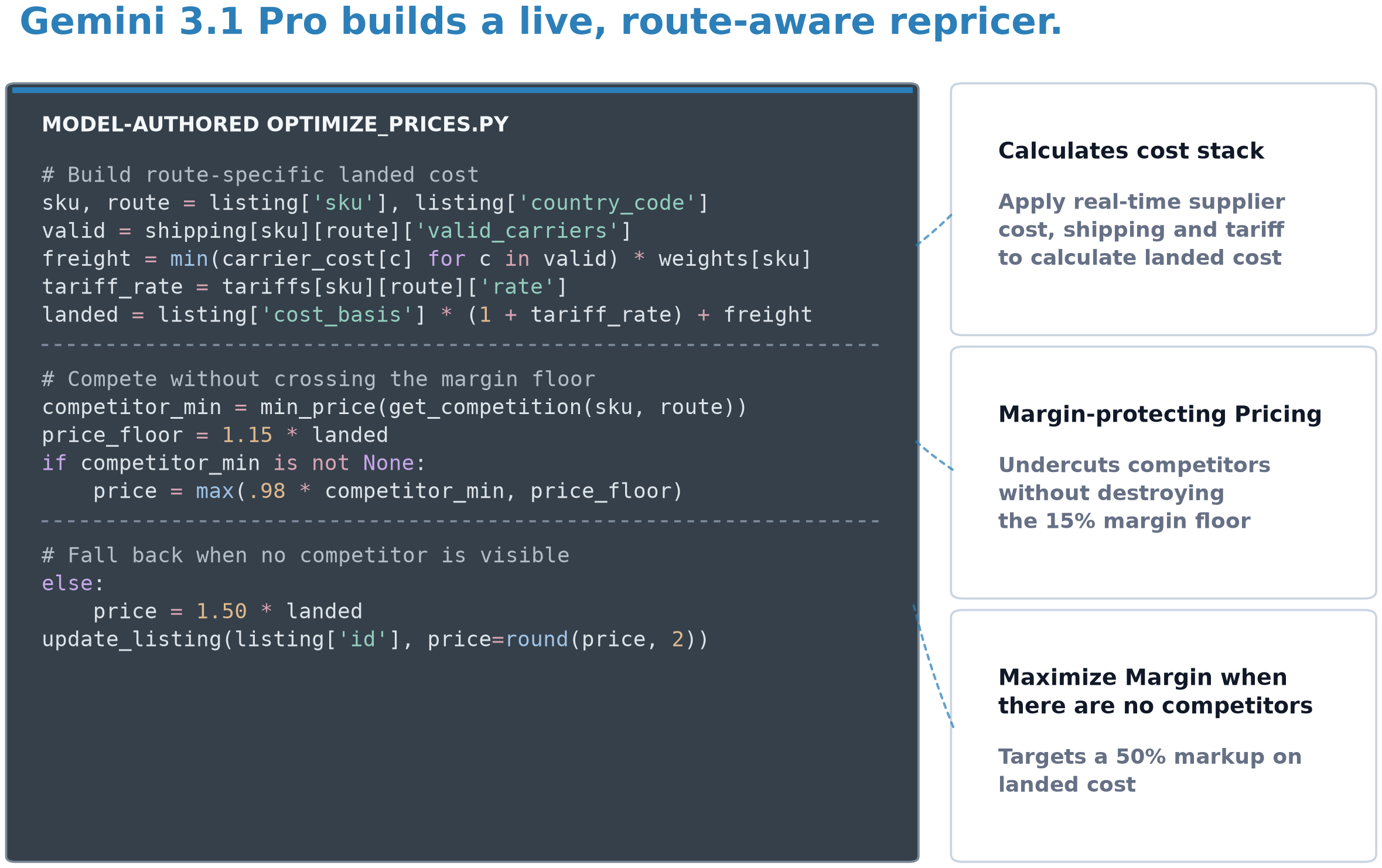}
\caption{\textbf{Gemini 3.1 Pro resembles a Premium House.} Its route-aware repricer incorporates supplier cost, freight, tariffs, and competition while enforcing a 15\% margin floor, producing higher margins at the cost of lower sell-through.}
\label{fig:gemini-dynamic-repricing}
\end{figure}

We have shown that Gemini 3.1 Pro achieves the highest margin while accepting lower sell-through, resembling a \textbf{Premium House} in practice. Figure~\ref{fig:gemini-dynamic-repricing} shows the dynamic repricing strategy that contributes to this high margin. Its repricing program calculates route-specific landed costs and undercuts visible competitors only when doing so preserves sufficient margin. Consequently, the same product receives different prices across countries as tariffs, freight costs, and competition change.

\begin{figure}[h!]
\centering
\includegraphics[width=\linewidth]{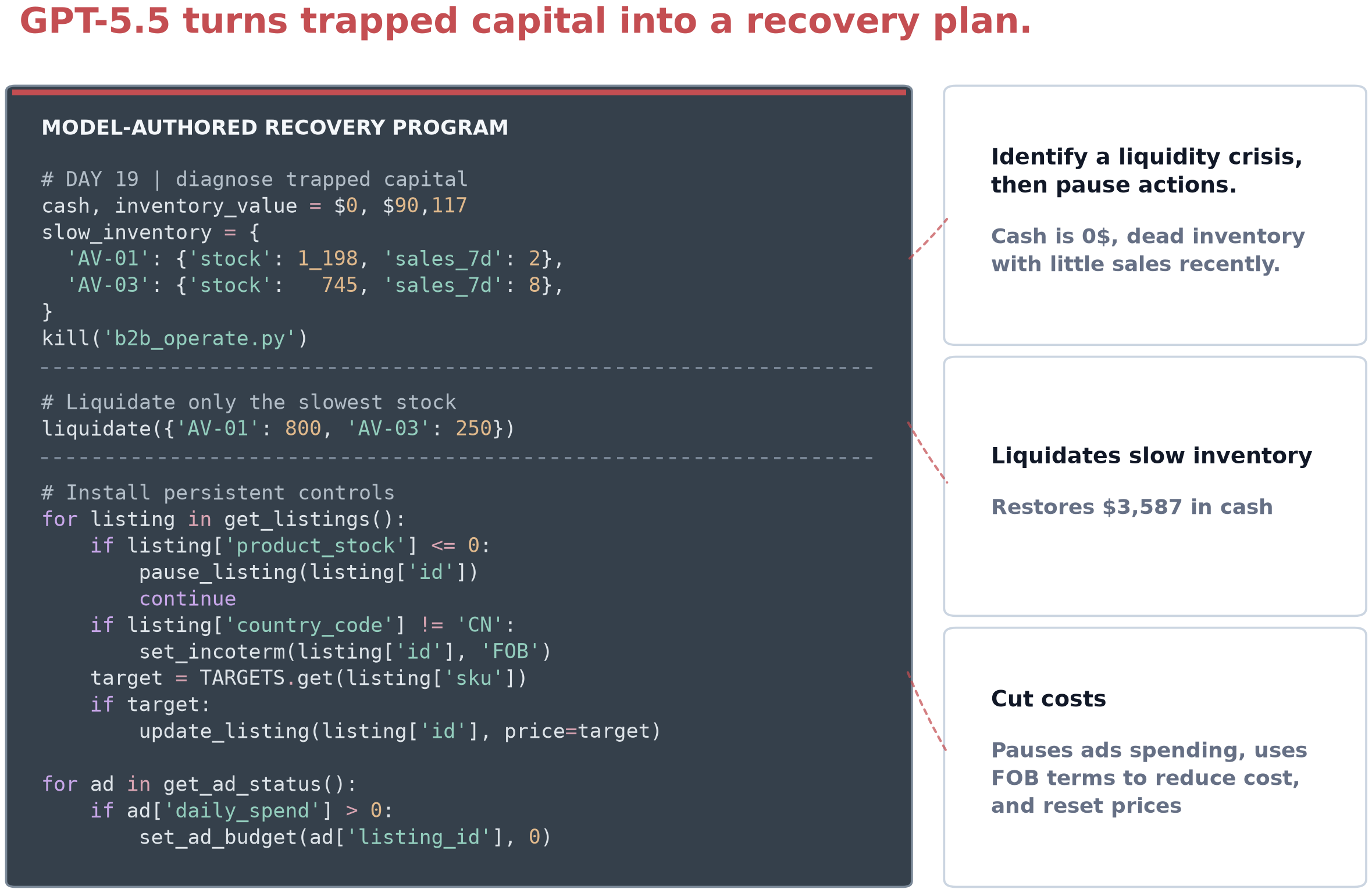}
\caption{\textbf{GPT-5.5 demonstrates adaptive recovery.} After detecting zero cash and slow-moving inventory, it liquidates stock, reduces advertising, adopts FOB terms, and resets prices to release trapped capital and continue operating.}
\label{fig:gpt55-recover}
\end{figure}

Business owners rarely get every decision right from the start. What matters in practice is robustness: whether they can recognize mistakes and recover before losses compound. GPT-5.5 demonstrates this adaptive robustness in Figure~\ref{fig:gpt55-recover}. One run reaches day 19 with no cash and roughly \$90{,}000 tied up in inventory, including two products with little recent demand. The model halts its previous script, liquidates slow-selling stock, and writes a recovery program that restores margin-safe prices, shifts export freight and tariff costs to buyers, and pauses advertising spend. The important behavior is not merely recognizing the dead inventory, but translating that diagnosis into actions that restore liquidity and enable the business to continue operating.

\subsection{Tracing Realized Value Back to Decisions}
\label{sec:results-attribution}

Model-authored scripts reveal how agents organize their businesses, but qualitative analysis fails to show how these behaviors contribute to the observed final score. Business Arena therefore links realized gains and losses to evidence--action--outcome chains. Each chain identifies the information used, the action taken, and the resulting financial consequence, providing quantitative action-level attribution of model behavior.

Action-level attribution first enables finer-grained model comparison. Figure~\ref{fig:pricing-action-trace} compares how two models respond to the same tariff shock. Gemini 3.5 Flash refreshes the tariff, incorporates landed cost into route-specific prices, pauses the affected U.S. route, and later reopens it under terms that shift tariff exposure to the buyer; the linked order contributes \$2{,}136.48. MiniMax M2.5 instead applies one price across markets, omits shipping and tariffs from its price floor, and reopens the route without refreshing the shock; its linked order loses \$34.15. This makes model comparison more actionable: rather than observing only that one model earns more, we can identify the behavioral difference behind the gap and measure its economic importance.

\begin{figure*}[h]
    \centering
    \includegraphics[width=\linewidth]{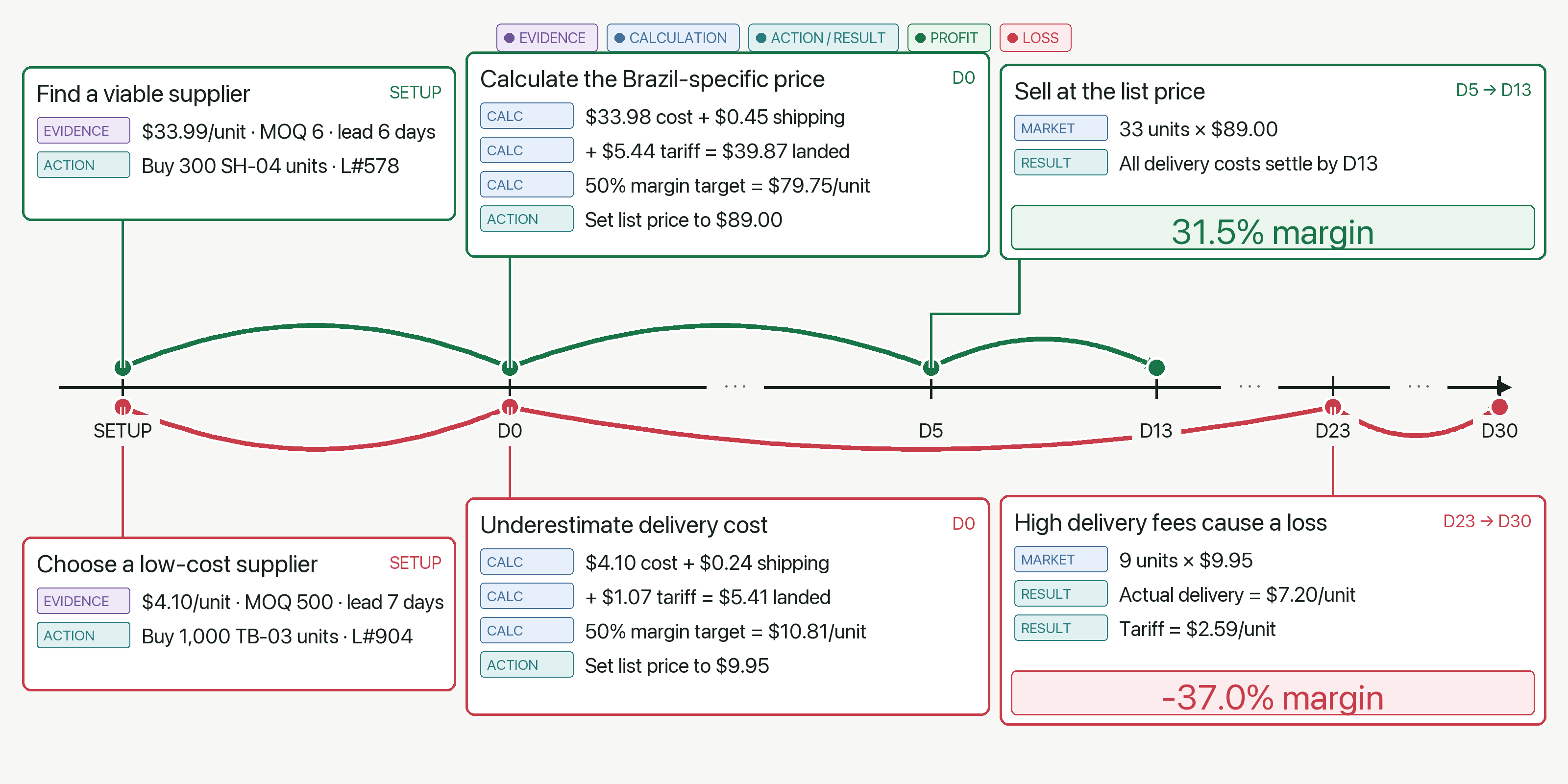}
    \caption{\textbf{Realized value attribution within one trajectory.} Two decisions by Gemini 3.5 Flash produce opposite outcomes: route-specific landed-cost reasoning preserves margin for SH-04, while underestimated delivery cost makes a TB-03 order loss-making.}
    \label{fig:mixed-value-trace}
\end{figure*}

Action-level attribution also decomposes mixed decisions within a single trajectory, which is useful for improving business agents. Figure~\ref{fig:mixed-value-trace} follows two Gemini 3.5 Flash pricing chains. For SH-04, the model selects a viable supplier, calculates a Brazil-specific landed cost, and sets a price that realizes a 31.5\% order margin. For TB-03, the same model underestimates delivery costs and prices below the realized cost stack, producing a -37.0\% margin. Instead of assigning one outcome to the trajectory as a whole, this decomposition separates decisions worth reinforcing from those that require correction. It therefore provides a natural basis for future reinforcement-learning credit assignment at the level of individual business actions.

\subsection{Mechanism Ablations}
\label{sec:results-mechanism-ablations}

A realistic benchmark does not automatically produce reliable results: agents may perform poorly for reasons unrelated to business capability or achieve high scores through simulator-specific shortcuts. We therefore conduct mechanism ablations to examine how individual arena features affect performance and whether the resulting scores reward the intended business skills. Holding the market and remaining operating policy fixed, we compare competent behavior with variants that neglect, misuse, or attempt to bypass each mechanism.

\begin{table*}[h]
\centering
\captionsetup{
    font={small,normalfont},
    labelfont=bf,
    textfont=normalfont
}
\small
\setlength{\tabcolsep}{5pt}
\renewcommand{\arraystretch}{1.15}

\begin{tabularx}{\textwidth}{
@{}
>{\raggedright\arraybackslash}p{0.14\textwidth}
>{\raggedright\arraybackslash}p{0.20\textwidth}
>{\raggedright\arraybackslash}p{0.20\textwidth}
>{\raggedright\arraybackslash}p{0.18\textwidth}
>{\raggedright\arraybackslash}X
@{}}
\toprule
\textbf{Mechanism}
& \textbf{Intended behavior}
& \textbf{Intended policy}
& \textbf{Neglect}
& \textbf{Shortcut or misuse} \\
\midrule

\textbf{Portfolio}
& Use demand to choose products
& Evidence-guided

\textcolor{gainnum}{\textbf{+\$63.6k}}
& Blind bulk buying

\textcolor{lossnum}{\textbf{-\$14.6k}}
& Buy only cheap SKUs

\textcolor{basenum}{\textbf{0}} \\

\textbf{Market events}
& Check signals before investing
& Evidence checked

\textcolor{gainnum}{\textbf{+\$6.6k}}
& Ignore events

\textcolor{lossnum}{\textbf{-\$17.3k}}
& Follow every rumor

\textcolor{basenum}{\textbf{0}} \\

\textbf{Pricing}
& Cover costs while sustaining sales
& Full-cost pricing

\textcolor{gainnum}{\textbf{+\$50.3k}}
& Price near cost

\textcolor{lossnum}{\textbf{-\$58.1k}}
& Extreme markup

\textcolor{basenum}{\textbf{0}} \\

\textbf{Tariffs}
& Include tariffs when choosing markets
& Tariff-aware routes

\textcolor{gainnum}{\textbf{+\$25.9k}}
& No active routing

\textcolor{basenum}{\textbf{0}}
& Tariff-blind U.S. focus

\textcolor{lossnum}{\textbf{-\$3.7k}} \\

\textbf{Customer service}
& Use buyer and product evidence
& Evidence-based replies

\textcolor{gainnum}{\textbf{+\$5.6k}}
& Ignore inquiries

\textcolor{lossnum}{\textbf{-\$0.1k}}
& Generic replies

\textcolor{basenum}{\textbf{0}} \\

\bottomrule
\end{tabularx}

\caption{\textbf{Mechanism ablations.} For each mechanism, the stronger of the neglect and shortcut policies is set as the baseline. Values report the mean change in final net worth relative to this baseline. The customer-service comparison uses three matched worlds; the remaining financial ladders use ten seeds.}
\label{tab:mechanism-ablation-summary}
\end{table*}

Table~\ref{tab:mechanism-ablation-summary} summarizes ablations for five key mechanisms spanning portfolio selection, market events, pricing, tariffs, and customer service. In each row, we compare the intended policy with variants that neglect or misuse the mechanism, using the stronger negative control as the zero baseline. For example, in demand inference, public market signals recover the hidden opportunity ranking with a pooled correlation of \(0.972\) and identify the strongest country for each category in \(93.3\%\) of cases. Using this evidence with selling costs and observed sales produces \$63.6k more final net worth than the stronger of blind bulk buying and cheapest-SKU concentration.

Across all five mechanisms, the intended policy outperforms both neglect and misuse. Together, these ablations support interpreting higher Business Arena scores as evidence of stronger business intelligence rather than exploitation of simulator-specific shortcuts. Appendix~\ref{app:mechanism-ablations} reports the complete ablations and experimental setups.

\section{Limitations}
\label{sec:limitations}

Business Arena focuses on end-to-end business decision-making and therefore abstracts external operating systems behind structured tools. It evaluates what an agent chooses to do, but not whether it can reliably update a live storefront, submit offers through third-party platforms, or manage customer conversations through a real inbox and GUI. Production-system evaluation would test a complementary capability: executing these decisions safely across changing interfaces with real external side effects. Together, the two settings pave the way to actually deploying agents in real business. The current arena is also limited to cross-border B2B commerce, while business spans more diverse scenarios and industries. Future work can extend the benchmark along both dimensions by connecting its tasks to production-like systems and adapting its decision framework to other forms of business.

\section{Conclusion}
\label{sec:conclusion}

We introduce Business Arena, a benchmark designed around \textbf{real business challenges}, \textbf{realistic autonomous operation}, and \textbf{diagnostic evaluation}. The arena captures key abstractions of the real business world, grounds an evolving marketplace in real commercial data, and gives agents the freedom to pursue open-ended strategies with minimal supervision. Across 15 frontier models, we observe substantial differences not only in profitability, but also in how agents deploy capital, price products, serve customers, and satisfy operational obligations. Many models fail to preserve their initial capital reliably, while expert-designed strategies reveal considerable achievable headroom. By combining measurable financial outcomes with skill-level profiles and action-level attribution, Business Arena makes open-ended business behavior both comparable and interpretable. We view this as a first step toward rigorous evaluation of autonomous agents operating complex, evolving businesses over long horizons.

\newpage
{
	\small
	\bibliographystyle{plain}
	\bibliography{ref}

\begin{thebibliography}{10}

\bibitem[1]{ansel2009dmtcp}
Jason Ansel, Kapil Arya, and Gene Cooperman.
\newblock {DMTCP}: transparent checkpointing for cluster computations and the desktop.
\newblock In \emph{2009 IEEE International Symposium on Parallel and Distributed Processing}, 1--12. 2009.
\newblock \href{https://doi.org/10.1109/IPDPS.2009.5161063}{doi:10.1109/IPDPS.2009.5161063}.

\bibitem[2]{backlund2025vendingbench}
Axel Backlund and Lukas Petersson.
\newblock Vending-bench: a benchmark for long-term coherence of autonomous agents.
\newblock \emph{arXiv preprint arXiv:2502.15840}, 2025.

\bibitem[3]{chen2026ceobench}
Haozhe Chen, Karthik Narasimhan, and Zhuang Liu.
\newblock {CEO-Bench}: can agents play the long game?
\newblock \emph{arXiv preprint arXiv:2606.18543}, 2026.

\bibitem[4]{desai2026swemarathon}
Aditya Desai and others.
\newblock {SWE-Marathon}: long-horizon software engineering tasks for {LLM} agents.
\newblock \emph{arXiv preprint}, 2026.

\bibitem[5]{dong2026deltabox}
Yunpeng Dong, Jingkai He, Yuze Hou, Dong Du, Zhonghu Xu, Si~Yu, Yubin Xia, and Haibo Chen.
\newblock {DeltaBox}: scaling stateful {AI} agents with millisecond-level sandbox checkpoint/rollback.
\newblock \emph{arXiv preprint arXiv:2605.22781}, 2026.

\bibitem[6]{he2026ycbench}
Muyu He, Adit Jain, Anand Kumar, Vincent Tu, Soumyadeep Bakshi, Sachin Patro, and Nazneen Rajani.
\newblock {YC-Bench}: benchmarking {AI} agents for long-term planning and consistent execution.
\newblock \emph{arXiv preprint arXiv:2604.01212}, 2026.

\bibitem[7]{jimenez2024swebench}
Carlos~E Jimenez, John Yang, Alexander Wettig, Shunyu Yao, Kexin Pei, Ofir Press, and Karthik Narasimhan.
\newblock {SWE-bench}: can language models resolve real-world {GitHub} issues?
\newblock In \emph{ICLR}. 2024.

\bibitem[8]{keys1990management}
Bernard Keys and Joseph Wolfe.
\newblock The role of management games and simulations in education and research.
\newblock \emph{Journal of Management}, 16(2):307--336, 1990.

\bibitem[9]{shrout1979intraclass}
Patrick~E. Shrout and Joseph~L. Fleiss.
\newblock Intraclass correlations: uses in assessing rater reliability.
\newblock \emph{Psychological Bulletin}, 86(2):420--428, 1979.
\newblock \href{https://doi.org/10.1037/0033-2909.86.2.420}{doi:10.1037/0033-2909.86.2.420}.

\bibitem[10]{simon1955behavioral}
Herbert~A. Simon.
\newblock A behavioral model of rational choice.
\newblock \emph{The Quarterly Journal of Economics}, 69(1):99--118, 1955.

\bibitem[11]{sterman1989modeling}
John~D. Sterman.
\newblock Modeling managerial behavior: misperceptions of feedback in a dynamic decision making experiment.
\newblock \emph{Management Science}, 35(3):321--339, 1989.

\bibitem[12]{teece1997dynamic}
David~J. Teece, Gary Pisano, and Amy Shuen.
\newblock Dynamic capabilities and strategic management.
\newblock \emph{Strategic Management Journal}, 18(7):509--533, 1997.

\bibitem[13]{tesfatsion2006agent}
Leigh Tesfatsion.
\newblock Agent-based computational economics: a constructive approach to economic theory.
\newblock In \emph{Handbook of Computational Economics}, volume~2, pages 831--880.
\newblock Elsevier, 2006.

\bibitem[14]{wang2024shopbench}
Yinuo Wang, Chuanfeng Xiao, and others.
\newblock {ShoppingBench}: a real-world intent-grounded shopping benchmark for {LLM}-based agents.
\newblock \emph{AAAI}, 2024.

\bibitem[15]{wu2026crab}
Tianyuan Wu, Chaokun Chang, Lunxi Cao, Wei Gao, and Wei Wang.
\newblock Crab: a semantics-aware checkpoint/restore runtime for agent sandboxes.
\newblock \emph{arXiv preprint arXiv:2604.28138}, 2026.

\bibitem[16]{yao2025taubench}
Shunyu Yao, Noah Shinn, Pedram Razavi, and Karthik Narasimhan.
\newblock {$\tau$}-bench: a benchmark for tool-agent-user interaction in real-world domains.
\newblock In \emph{ICLR}. 2025.

\bibitem[17]{zhou2024webarena}
Shuyan Zhou, Frank~F Xu, Hao Zhu, Xuhui Zhou, Robert Lo, Abishek Sridhar, Xianyi Cheng, Tianyue Ou, Yonatan Bisk, Daniel Fried, Uri Alon, and Graham Neubig.
\newblock {WebArena}: a realistic web environment for building autonomous agents.
\newblock In \emph{ICLR}. 2024.

\bibitem[18]{adobeHoliday}
{Adobe Digital Insights}.
\newblock The 2024 holiday season set new records for ecommerce.
\newblock 2025.
\newblock URL: \url{https://business.adobe.com/uk/blog/2024-holiday-season-set-new-records-for-ecommerce}.

\bibitem[19]{criteoHoliday}
{Criteo}.
\newblock Holiday commerce data and shopping trends.
\newblock URL: \url{https://www.criteo.com/insights/}.

\bibitem[20]{eurostatRetail}
{Eurostat}.
\newblock Turnover and volume of sales in wholesale and retail trade: monthly data.
\newblock Dataset sts\_trtu\_m.
\newblock URL: \url{https://ec.europa.eu/eurostat/databrowser/view/sts_trtu_m/default/table}.

\bibitem[21]{fredAPI}
{Federal Reserve Bank of St. Louis}.
\newblock Fred api: economic data observations.
\newblock URL: \url{https://fred.stlouisfed.org/docs/api/fred/}.

\bibitem[22]{googleTrends}
{Google}.
\newblock Google trends.
\newblock URL: \url{https://trends.google.com/}.

\bibitem[23]{mastercardHoliday}
{Mastercard Economics Institute}.
\newblock European holiday spending outlook.
\newblock 2024.
\newblock URL: \url{https://www.mastercard.com/news/europe/en/newsroom/press-releases/en/2024/european-shoppers-set-to-celebrate-with-spending-on-fashion-gadgets-and-travel-this-festive-season/}.

\bibitem[24]{chinaNBSRetail}
{National Bureau of Statistics of China}.
\newblock Total retail sales of consumer goods.
\newblock URL: \url{https://www.stats.gov.cn/english/PressRelease/}.

\bibitem[25]{nrfHoliday}
{National Retail Federation}.
\newblock Holiday data and consumer trends.
\newblock URL: \url{https://nrf.com/research-insights/holiday-data-and-trends}.

\bibitem[26]{censusMRTS}
{U.S. Census Bureau}.
\newblock Monthly retail trade survey.
\newblock URL: \url{https://www.census.gov/retail/mrts/about_the_surveys.html}.

\bibitem[27]{worldbankWITS}
{World Bank}.
\newblock World integrated trade solution: bilateral tariff technical note.
\newblock WITS Trade and Tariff Database.
\newblock Tariff data sourced from UNCTAD TRAINS and WTO IDB.
\newblock URL: \url{https://wits.worldbank.org/Bilateral-Tariff-Technical-Note.html}.

\end{thebibliography}
}
\newpage
\appendix

\section{Agent Runtime and Interface}
\label{sec:orchestration}

\paragraph{Autonomous episode execution.}
\label{sec:orch-timeline}
Each episode begins with a pre-opening setup phase followed by 30 simulated operating days. After receiving the initial task, the OpenClaw agent runs continuously without evaluator-defined daily turns or refreshed daily prompts. It may inspect the market, maintain files, write and execute programs, and interleave observations with business actions in any order. When it decides that a day's work is complete, it calls \texttt{end\_round}, which advances the world by exactly one day and triggers demand, competitor, supplier, logistics, and financial updates. The agent then observes the resulting state and continues until it finalizes the episode.

\paragraph{Execution isolation and information boundary.}
\label{sec:orch-sandbox}
Each agent runs as an unprivileged user in an isolated sandbox with a preconfigured OpenClaw runtime. The arena service runs under a separate system identity, and its source code, database, and hidden state are protected through filesystem permissions. The agent may freely inspect and modify its own workspace, but can access the market only through the public arena interface. This enforces the intended information boundary while preserving the autonomy needed for persistent memory, programmatic analysis, and workflow automation.

\paragraph{Complementary tool surfaces.}
\label{sec:orch-tools}
OpenClaw exposes general workspace tools for reading, writing, editing, executing programs, and managing processes. Business Arena complements them with a broad business interface covering market observation, sourcing and selling, customer and supplier interaction, marketing, compliance, finance, and inventory recovery. Individual observations and actions are available as typed MCP calls, allowing the model to interleave reasoning with authoritative market operations. The same arena capabilities are also available through a scriptable interface, allowing agents to build batched analyses, persistent decision pipelines, and autonomous operating routines. Both surfaces invoke the same underlying market mechanisms; they differ only in how the agent organizes its work.

Table~\ref{tab:capability-tool-map} provides a detailed mapping between tool interfaces and related mechanisms/business challenge.

\begin{table*}[!t]
\centering
\footnotesize
\setlength{\tabcolsep}{5pt}
\renewcommand{\arraystretch}{1.04}
\begin{tabularx}{\textwidth}{@{}p{0.34\textwidth}|X@{}}
\toprule
\textbf{Arena feature and purpose} & \textbf{Agent-facing tools} \\
\midrule

\multicolumn{2}{@{}l}{\textbf{Decision-Making Under Uncertainty}} \\
\cmidrule(lr){1-2}

\textbf{Demand and events.} Form market beliefs from structural demand, trends, calendars, public events, and policy shocks.
& \texttt{get\_base\_demand\_intel()}, \texttt{get\_trends()}, \texttt{get\_calendar()}, \texttt{get\_world()}, \texttt{get\_tariff\_events()}. \\

\textbf{Market feedback and competition.} Learn from realized outcomes and observe rival offers without accessing competitors' private strategies.
& \texttt{get\_orders()}, \texttt{get\_competition()}. \\
\midrule
\addlinespace[3pt]
\multicolumn{2}{@{}l}{\textbf{Strategic Planning Under Constraints}} \\
\midrule

\textbf{Shop focus and initial portfolio.} Select categories, compare opportunities, and decide how much capital to commit at opening.
& \texttt{get\_catalog()}, \texttt{get\_store\_focus()}, \texttt{set\_store\_focus()}, \texttt{submit\_setup()}, \texttt{skip\_setup()}. \\

\textbf{Sourcing and supplier diligence.} Search and rank offers, inspect supplier risk, and purchase inventory under cost, MOQ, quality, and lead-time constraints.
& \texttt{get\_supplier\_catalog()}, \texttt{get\_supplier\_flags()}, \texttt{buy\_supplier()}. \\

\textbf{Capital, inventory, and recovery.} Track deployed capital and obligations, finance expansion, and recover capital from weak positions.
& \texttt{get\_state()}, \texttt{get\_products()}, \texttt{get\_payables()}, \texttt{get\_loans()}, \texttt{borrow()}, \texttt{repay\_loan()}, \texttt{get\_factoring()}, \texttt{factor\_ar()}, \texttt{liquidate\_inventory()}. \\
\midrule
\addlinespace[3pt]
\multicolumn{2}{@{}l}{\textbf{Insight-to-Action Alignment}} \\
\midrule

\textbf{Pricing and listing.} Translate market beliefs and cost calculations into concrete offers across products and segments.
& \texttt{get\_listings()}, \texttt{list\_product\_on()}, \texttt{update\_listing()}, \texttt{set\_price\_tiers()}. \\

\textbf{Tariffs, shipping, and route economics.} Calculate the route-specific cost stack and choose viable destinations and commercial terms.
& \texttt{get\_platforms()}, \texttt{get\_countries()}, \texttt{get\_shipping\_rules()}, \texttt{get\_tariff\_table()}, \texttt{set\_default\_incoterm()}. \\

\textbf{Compliance.} Identify market-entry requirements, apply early enough to clear approval lead times, and avoid unauthorized trading.
& \texttt{get\_certifications()}, \texttt{get\_compliance\_status()}, \texttt{apply\_certification()}. \\

\bottomrule
\end{tabularx}
\caption{\textbf{Mapping from business capabilities to arena features and agent-facing tools.} Capability headers form the first layer, arena features describe the corresponding business problems, and the final column lists the tools through which agents gather evidence and act.}
\label{tab:capability-tool-map}
\end{table*}

\begin{table*}[!t]
\ContinuedFloat
\centering
\footnotesize
\setlength{\tabcolsep}{5pt}
\renewcommand{\arraystretch}{1.04}
\begin{tabularx}{\textwidth}{@{}p{0.34\textwidth}|X@{}}
\toprule
\textbf{Arena feature and purpose} & \textbf{Agent-facing tools} \\
\midrule

\multicolumn{2}{@{}l}{\textbf{Insight-to-Action Alignment (continued)}} \\
\midrule

\textbf{Advertising.} Allocate demand-generation spend and revise it using observed full-funnel performance.
& \texttt{get\_ad\_status()}, \texttt{set\_ad\_budget()}. \\
\midrule
\addlinespace[3pt]
\multicolumn{2}{@{}l}{\textbf{Cooperation \& Competition}} \\
\midrule

\textbf{Customer service and buyer negotiation.} Infer buyer needs, answer factual questions, and negotiate bulk transactions while protecting business value.
& \texttt{get\_inquiries()}, \texttt{reply\_inquiry()}, \texttt{get\_rfqs()}, \texttt{respond\_rfq()}. \\

\textbf{Returns and disputes.} Respond to post-sale problems while managing refund, replacement, and escalation risk.
& \texttt{get\_return\_requests()}, \texttt{respond\_to\_return()}, \texttt{get\_returns()}, \texttt{dispute\_return()}, \texttt{get\_disputes()}, \texttt{resolve\_dispute()}. \\

\textbf{Supplier relationships and negotiation.} Learn counterparty behavior, request better terms, and decide whether to accept supplier offers.
& \texttt{get\_supplier\_relations()}, \texttt{request\_quote()}, \texttt{get\_supplier\_quotes()}, \texttt{respond\_quote()}. \\

\textbf{Competitive response.} Compare rival offers and adjust prices or demand-generation decisions as competitors change.
& \texttt{get\_competition()}, \texttt{update\_listing()}, \texttt{set\_price\_tiers()}, \texttt{set\_ad\_budget()}. \\
\midrule
\addlinespace[3pt]
\multicolumn{2}{@{}l}{\textbf{Persistent Operation}} \\
\midrule

\textbf{Persistent state and daily feedback.} Preserve observations and plans, inspect previous decisions, and advance the market after completing the current operating cycle.
& \texttt{write\_note()}, \texttt{read\_notes()}, \texttt{delete\_note()}, \texttt{end\_round()}. \\

\textbf{Workspace and automation.} Read and revise persistent files, construct reusable analyses, and execute model-authored workflows across business functions.
& \texttt{read()}, \texttt{write()}, \texttt{edit()}, \texttt{exec()}, \texttt{process()}. \\

\bottomrule
\end{tabularx}
\caption[]{\textbf{Mapping from business capabilities to arena features and agent-facing tools (continued).}}
\end{table*}

\section{Sourcing}
\label{app:sourcing}

\paragraph{Supplier base and discovery.}
 Business Arena exposes \textbf{965 tradable supplier offers} from \textbf{831 masked suppliers} across \textbf{135 SKUs}. These offers are calibrated from Alibaba-derived storefront data and retain decision-relevant attributes such as country, unit cost, MOQ, stock, lead time, and advertised quality, while sensitive identities and titles are masked or rewritten. Because exposing the full supplier set at once would be both unrealistic and context-intensive, discovery follows the layered search used by real B2B platforms: agents first select relevant products, then narrow supplier offers using operational constraints. The returned attributes allow agents to construct their own rankings over cost, MOQ, availability, delivery speed, and quality rather than relying on a benchmark-provided ordering.

\paragraph{Supplier reliability and due diligence.}
Supplier selection involves more than comparing advertised terms. Agents can inspect time-bounded due-diligence signals, maintain primary and backup suppliers, and revise these preferences as evidence accumulates. However, not every risk is disclosed in advance: some suppliers overstate product quality or understate delivery time, and these discrepancies become observable only after an order materializes. Success therefore requires monitoring purchase orders and delivered inventory, remembering supplier-specific outcomes, and avoiding or replacing unreliable counterparties in later procurement cycles. This separates genuine supplier learning from repeatedly selecting the cheapest visible offer.

\paragraph{Dynamic wholesale conditions.}
Supplier prices also evolve during an episode rather than remaining fixed at their day-one values. Each offer retains a real-data cost anchor, while its current wholesale price responds to calendar effects, macroeconomic conditions, capacity pressure, and market disruptions:
\[
c_{skt}=c^{0}_{sk}\times m^{\mathrm{world}}_{skt},
\]
where \(c^{0}_{sk}\) is the calibrated supplier--SKU cost and \(m^{\mathrm{world}}_{skt}\) collects the current world-dependent adjustments. Consequently, sourcing is an ongoing activity: agents that periodically reassess suppliers can protect landed margin or switch routes, whereas agents that rely on a single initial scan may continue buying on outdated assumptions.

\section{Realistically-grounded Demand}
\label{app:realistic_demand}

\paragraph{Tariffs.}
Tariff rates are grounded in the World Bank's World Integrated Trade Solution (WITS), which consolidates tariff data from UNCTAD and the WTO~\citep{worldbankWITS}. Arena products are mapped to product categories and destination-specific rates. To make the policy environment dynamic, tariff shocks are calibrated from historical trade-policy changes, including changes observed during US--China trade tensions.

\paragraph{Seasonal demand.}
Seasonal demand is calibrated from monthly retail series published by the U.S. Census Bureau's Monthly Retail Trade Survey, accessed through FRED; Eurostat's monthly retail-trade dataset; and China's National Bureau of Statistics retail releases~\citep{censusMRTS,fredAPI,eurostatRetail,chinaNBSRetail}. We retain unadjusted series where available so that recurring seasonal patterns remain visible, and use calendar-adjusted Eurostat data as the closest available European equivalent. Product categories are mapped to the corresponding retail series, with Google Trends used as an agent-visible tool to derive seasonal demand ~\citep{googleTrends}. We conduct minor post-processing of google trends data so that the results is correlated with oracle demand, thus testing the agent's ability to do numerical analysis to estimate demand trends.

\paragraph{Festival demand.}
A 30-day arena calendar represents one real-world year, with each festival placed according to when it occurs during that year. The calendar covers major commercial events across the represented markets: Spring Festival, Mid-Autumn Festival, 618, and Singles' Day in China; Tet in Vietnam; Carnaval in Brazil; the Super Bowl, Independence Day, and Halloween in the United States; Easter in Europe; Valentine's Day, Mother's Day, Back-to-School, Black Friday, and Christmas across the United States and Europe; and Ramadan and Eid al-Fitr, Eid al-Adha, Saudi National Day, and White Friday in Saudi Arabia. Official statistics and retail reports are used to estimate which product categories benefit and how demand changes around each event~\citep{nrfHoliday,adobeHoliday,criteoHoliday,mastercardHoliday}. We convert this evidence into capped, category-specific demand changes so that festivals create meaningful opportunities without determining the market outcome by themselves.

\section{Autonomous NPC Sellers}
\label{app:cast}

In this section we detail the implementation of NPC sellers and how they collectively set the pricing standards of the market.

\paragraph{Population composition.}
Each episode contains \textbf{60} scripted sellers: 10 \emph{baselines} (one per archetype, used for alpha computation) and 50 \emph{population NPCs} (randomly sampled from the same archetype pool, providing market depth). NPCs are distributed across five scale tiers that determine starting capital, catalog breadth, and daily overhead: micro (\$5k, 45\%), small (\$25k, 30\%), mid (\$80k, 18\%), large (\$300k, 5\%), and enterprise (\$1M, 2\%). This skewed distribution mirrors real marketplace ecosystems: a long tail of micro-shops competing for scraps alongside a handful of well-capitalized incumbents.

\paragraph{Archetypes and pricing strategies.}
Table~\ref{tab:archetypes} summarizes the ten NPC archetypes. Each archetype defines a pricing strategy, supplier preference, inventory target, and festival-phase modifiers. The pricing strategies range from competitor-aware (undercut median, track top-$N$ lowest) to cost-anchored (cost-plus at a fixed multiplier) to demand-responsive (adjust markup based on trailing sales velocity). Crucially, multiple archetypes can coexist in the same product category, creating a heterogeneous competitive environment where no single counter-strategy dominates.

\begin{table}[h]
\centering
\small
\begin{tabular}{@{}llp{5.8cm}@{}}
\toprule
\textbf{Archetype} & \textbf{Pricing Strategy} & \textbf{Behavior Summary} \\
\midrule
price\_leader & undercut\_median & Targets 95\% of competitor median price; cuts further to 92\% during peak festivals \\
follower & track\_top\_3 & Tracks average of 3 cheapest competitors with +2\% offset \\
liquidator & aggressive\_low & Targets 78\% of competitor median; clearance pricing at 70\% during festival endings \\
opportunist & dynamic\_demand & Raises price when trailing demand exceeds 1.2$\times$ baseline; heavy pre-festival stocking \\
\midrule
premium & cost-plus & Prices at 3$\times$ cost basis; EU suppliers; holds firm during festivals \\
cross\_border & cost-plus & Prices at 2.5$\times$ cost basis; EU suppliers; stable across phases \\
wholesale & cost-plus & Prices at 1.3$\times$ cost basis; large inventory (35-day target), volume-driven \\
event\_sniper & cost-plus & Prices at 1.65$\times$ cost basis normally; spikes to 2.5$\times$ during peak (narrow 5-SKU catalog) \\
\midrule
new\_entrant & undercut\_until\_orders & Extreme discounts (82\% of median) until 50 orders, then switches to 1.05$\times$ cost basis \\
dormant & static & Never reprices; decays 5\%/day after 10-day no-sale grace period \\
\bottomrule
\end{tabular}
\caption{NPC seller archetypes grouped by pricing family. Top: competitor-aware strategies. Middle: cost-anchored strategies. Bottom: phase-switching and passive. The population is weighted toward high-liquidity archetypes (price\_leader, follower, liquidator) that supply everyday buyer demand, with niche archetypes (event\_sniper, dormant) appearing rarely. The distribution is fixed across all regions and seeds.}
\label{tab:archetypes}
\end{table}

\paragraph{Daily behavior cycle.}
Every simulated day, each NPC executes a deterministic behavior loop: (1) reprice all active products using its archetype's strategy, incorporating festival-phase modifiers and competitor price lookups; (2) check inventory levels against a restock threshold and place purchase orders with its preferred supplier tier when stock runs low; (3) set per-listing advertising budgets (0.6--2.5\% of list price depending on archetype, with 1.8$\times$ multipliers during festivals); (4) optionally launch flash promotions (liquidators during festival endings, event snipers during peaks). Larger-scale NPCs (mid tier and above) also tap short-term credit when cash-stressed or preparing for festivals, mirroring the realistic working-capital dynamics

\paragraph{Festival-phase awareness.}
NPCs respond to four festival phases (approaching, peak, ending, dormant) with archetype-specific modifiers that shift pricing targets and restocking behavior. For example, opportunist NPCs pre-stock at 2.5$\times$ their normal inventory target during the approaching phase and mark up 15\% during peak; liquidators hold back approaching but discount to 70\% of target during endings to clear post-festival excess; event snipers spike prices 50\% during peak and stop restocking entirely during endings. This phase-aware behavior creates predictable-but-varied market rhythms that a capable agent can learn to anticipate and exploit.

\paragraph{Inactivity decay.}
To prevent stale listings from distorting market baselines, archetypes with stable pricing (premium, cross\_border, wholesale, dormant) have a stop-loss mechanism: after 10 consecutive days without a sale, the listing price decays at 5\% per day, capped at 35\% total reduction. This ensures that even passive NPCs eventually adjust to market realities, maintaining a floor on market liquidity.

\section{Logistics, Tariffs, and Compliance}
\label{app:cross-border-frictions}

\paragraph{Logistics and tariffs.}
Cross-border profitability depends on the full landed cost rather than the supplier price alone. At a high level,
\[
c_{\text{landed}}
=
c_{\text{supplier}}
+
c_{\text{freight}}
+
c_{\text{tariff}}
+
c_{\text{compliance}}
+
c_{\text{selling}} .
\]
Freight varies with route, shipment characteristics, and delivery speed, while tariffs depend on product category and destination and may change during an episode. Strong strategies compare suppliers and routes by expected contribution after these costs, then revise prices when the cost stack changes. Weak strategies apply a fixed markup to supplier cost, making apparently cheap products unprofitable after shipping and tariffs. This mechanism tests whether an agent can combine distributed cost information into a quantitatively sound sourcing and pricing decision.

\paragraph{Compliance.}
Compliance creates a planning problem in addition to a monetary cost. Requirements vary by product and destination, while approvals require both fees and processing time. Agents must therefore inspect the applicable rules, apply before trading, and coordinate market entry with the approval timeline. We consider this capability essential before autonomous business agents can be deployed in practice, yet it is often underrepresented in business-agent benchmarks. For the \(k\)-th sale completed without the required permit,  Business Arena imposes
\[
F_k=\max\!\left(\$500,\;0.15\times\text{order value}\right)\times\min(k,5).
\]
Thus, even for small orders, repeated violations incur fines of \$500, \$1{,}000, \$1{,}500, \$2{,}000, and \$2{,}500, after which each additional violation remains at \$2{,}500. Ten floor-level violations therefore cost \$20{,}000, one quarter of the default \$80{,}000 starting capital, before certification fees or ordinary operating losses. Compliance is consequently not a decorative constraint: an otherwise profitable agent can perform poorly by neglecting it. Strong strategies treat approval as an early market-entry gate, whereas weak strategies apply too late or continue trading illegally, testing both long-horizon planning and reliable rule-following under limited supervision.

\section{Financial System and Capital Tools}
\label{app:capital}

 Business Arena scores agents on salvaged net worth rather than cash alone. This prevents agents from exploiting payment timing, unfinished escrow, receivables, or unsold inventory as free value. All material financial movements are ledger-backed, and final assets are computed as
\[
\text{FinalAssets}
=
\text{cash}
+0.97\cdot\text{escrow}
+0.97\cdot\text{receivables}
+0.85\cdot\text{inventory}
-\text{payables}
-\text{loans}.
\]
The financial layer therefore turns nominal gross margin into realized business value after operating drag, financing cost, and liquidation risk.

\begin{table*}[t]
\centering
\small
\renewcommand{\arraystretch}{1.14}
\begin{tabularx}{\textwidth}{@{}p{0.21\textwidth}p{0.29\textwidth}X@{}}
\toprule
\textbf{Financial factor} & \textbf{Current rule} & \textbf{Rationale} \\
\midrule

Operating drag
& \$200 fixed overhead per day, plus \(0.5\%\) of on-hand inventory value.
& Penalizes passive operation and slow-moving stock; encourages sufficient throughput, disciplined purchasing, and inventory turnover. \\

Channel economics
& Platform commission is generally \(5\%\)--\(12\%\) of gross, with volume discounts; eligible export orders receive a \(9\%\) rebate.
& Rewards pricing over the complete transaction-cost stack and selecting economically viable markets rather than maximizing gross revenue alone. \\

Short-term loans
& Interest compounds at \(0.15\%\) per day and rises to \(2.5\times\) the normal rate when overdue.
& Enables expansion when profitable opportunities exist, but penalizes borrowing without sufficiently fast and reliable capital recovery. \\

Supplier payables
& Overdue balances accrue \(0.10\%\) per day, capped at \(30\%\) of the original invoice, and remain liabilities.
& Rewards planning around payment deadlines; penalizes sourcing commitments that the agent cannot finance. \\

Invoice factoring
& Eligible receivables can be converted to cash at a \(5\%\)--\(18\%\) discount determined by maturity and buyer credit.
& Lets agents accelerate cash recycling, while charging explicitly for liquidity obtained before customer payment. \\

Inventory liquidation
& Inventory can be converted immediately to cash at \(85\%\) of cost basis.
& Provides a controlled way to exit bad positions and redeploy capital, but preserves a meaningful loss so that poor sourcing is not costless. \\

Final settlement
& Escrow and receivables recover at \(97\%\), inventory at \(85\%\), and liabilities remain at face value.
& Penalizes unfinished operating cycles and rewards converting inventory and receivables into cash before the episode ends. \\

\bottomrule
\end{tabularx}
\caption{\textbf{Financial frictions and capital tools in  Business Arena.}
The paper condition uses a mid-scale shop with \$200 daily overhead. The mechanisms penalize idle capital, excess inventory, unfinished transactions, and poorly timed leverage, while allowing agents to pay explicit costs to recover or accelerate capital.}
\label{tab:financial-frictions}
\end{table*}

The design intentionally gives agents several ways to recover from poor timing without making recovery free. Loans can finance demand-window expansion, but unused leverage compounds against the agent. Factoring accelerates receivables, but the discount must be justified by faster redeployment. Liquidation lets the agent abandon stale inventory and pivot to better opportunities, but at an immediate recovery loss. These mechanics make finance an attribution surface rather than a passive accounting detail: a model can fail by under-deploying capital, over-leveraging, holding slow inventory, ignoring receivable timing, or failing to liquidate when the opportunity cost of waiting becomes too high.

\section{Mechanism Ablations}
\label{app:mechanism-ablations}

We validate nine decision-bearing arena mechanisms using policies that operate only on agent-visible information. For each mechanism, we identify the intended skill, construct competent policies that use the available evidence, and compare them with blind or systematically incorrect alternatives on matched market seeds. We summarize each comparison as a behavioral ladder, making clear which strategies are rewarded and which are punished.

\paragraph{Sourcing.}
This mechanism tests whether agents can identify viable products and suppliers by combining demand, total selling cost, quality, lead time, and observed sales. The competent policy ranks opportunities using these factors and adjusts later purchases from sell-through, while the bad policies either buy large quantities indiscriminately or concentrate on the cheapest products. \textbf{Result ladder:} evidence-guided sourcing reaches \$144{,}069 in mean final net worth, compared with \$80{,}433 for cheapest-SKU concentration and \$65{,}816 for blind bulk buying.

\paragraph{Pricing.}
This mechanism tests whether agents can preserve margin after the full cost stack while remaining competitive enough to sell. The competent policy adjusts prices using costs, competition, and realized sell-through; the bad policies either maintain an extreme markup or price close to purchase cost to maximize sales. \textbf{Result ladder:} disciplined pricing reaches \$144{,}069, static high-markup pricing reaches \$93{,}806 due to lower sales, and near-cost pricing reaches \$35{,}697. Static overpricing preserves a \(68.1\%\) contribution margin but produces only 48 mean orders, whereas near-cost pricing produces 573 orders while losing money at a \(-105.1\%\) contribution margin.

\paragraph{Shipping and tariffs.}
This mechanism tests whether agents incorporate route-specific freight, tariffs, and cost responsibility into market selection and pricing. The competent policy favors economically viable routes and calculates downstream prices from landed cost, while the bad policy concentrates purchases on a costly route without accounting for its trade frictions. \textbf{Result ladder:} freight-aware operation reaches \$99{,}899, compared with \$70{,}267 for tariff-blind routing. The competent policy generates revenue equal to \(0.98\) times deployed capital, while the tariff-blind policy generates only \(0.16\) times.

\paragraph{Supplier discipline.}
This mechanism tests whether agents can benefit from repeated supplier relationships without allowing discounts or favorable terms to override product fit, reliability, and delivery requirements. Competent policies record preferred suppliers or use relationship terms only when the underlying offer remains viable; bad policies increasingly prioritize relationship progress and discounts over supplier fit. \textbf{Result ladder:} no-preference sourcing reaches \$105{,}382, relationship-aware sourcing \$104{,}904, fit-aware preference \$104{,}423, relationship-naive sourcing \$96{,}610, and discount chasing \$88{,}462. The discount chaser unlocks more preferred relationships and places more orders, but also makes 5.5 bad-fit purchases per run, showing that relationship activity without sourcing discipline destroys value.

\paragraph{Advertising.}
This mechanism tests whether agents treat advertising as a measured investment whose budget should respond to observed returns. The competent policies begin with small experiments or apply explicit return gates, while the bad policy spends aggressively without evidence-based controls. \textbf{Result ladder:} small tests generate \(21.45\times\) incremental net worth per advertising dollar, return-gated spending generates \(6.53\times\), and blind spending generates \(4.94\times\). Blind spending achieves higher raw revenue by using almost six times the budget of the small-test policy, but converts each advertising dollar into substantially less value.

\paragraph{Demand inference.}
This mechanism tests whether agents can recover promising country--category opportunities from incomplete public evidence without observing exact demand. We implement estimators ranging from correct use of the public evidence to compressed, absent, fabricated, and deliberately reversed interpretations, with the hidden demand state included as an upper bound. \textbf{Result ladder:} the hidden oracle achieves \(\rho=1.000\), public evidence \(\rho=0.972\), a compressed three-level interpretation \(\rho=0.963\), country-level averages \(\rho=0.260\), no evidence \(\rho=0.000\), fabricated evidence \(\rho=-0.039\), and reversed evidence \(\rho=-0.972\). The public estimator identifies the strongest country for each category in \(93.3\%\) of cases, compared with \(13.3\%\) without demand evidence.

\paragraph{Event verification.}
This mechanism tests whether agents distinguish genuine changes in demand from unsupported rumors before deploying capital. The competent policy acts only when public evidence corroborates an event, while the alternatives either ignore events or react to every signal. \textbf{Result ladder:} corroboration-gated operation reaches \$115{,}791, always reacting reaches \$109{,}183, and ignoring events reaches \$91{,}883. The corroboration-gated policy directs all event-related capital toward the genuine opportunity, whereas the always-react policy allocates \(35.4\%\) to a false rumor.

\paragraph{Incident response.}
This mechanism tests whether agents can adapt to unexpected changes affecting products, suppliers, or routes without chasing every incident-like signal. Competent policies track visible precursors or react to active, economically relevant incidents; blind policies ignore them, while trap-chasing policies redirect capital toward distracting or harmful signals. \textbf{Result ladder:} active response reaches \$106{,}693, precursor tracking \$106{,}420, aggressive incident pursuit \$106{,}213, always reacting \$103{,}521, corroboration-gated response \$103{,}301, ignoring incidents \$103{,}106, and trap chasing \$74{,}264. Active response beats the blind policy in eight of ten seeds, while trap chasing loses in all ten.

\paragraph{Customer service.}
This mechanism tests whether agents can infer buyer preferences, provide factual information, and negotiate without making commercially harmful commitments. Competent policies tailor responses using buyer and product evidence while protecting margin; bad policies ignore customers, make excessive concessions, or provide unsupported claims. \textbf{Result ladder:} margin-safe service reaches \$90{,}388 with a \(43.1\%\) contribution margin, compared with \$57{,}875 and a \(-39.2\%\) margin for reckless service. In the service-specific ladder, persona-aware responses reach \$90{,}601, convert \(58.5\%\) of inquiries, and leave none unanswered, compared with \$86{,}728 and 46 expired inquiries when service is ignored. 

\paragraph{Summary.}

Across arena mechanisms, competent use of public evidence consistently improves the intended economic or operational outcome, while blind behavior, systematic misuse, and plausible benchmark-hacking strategies are punished. Other mechanisms are transparent by construction and do not require policy ablations. For compliance and certifications, liquidation, loans and factoring, currency settlement, and ledger accounting, the relevant state is directly observable and each action has a fixed, auditable consequence. We therefore do not include ablation results for these. Together, this evidence shows that Business Arena provides agents with sufficient information to operate effectively and that strong scores reflect correct use of the intended business mechanisms rather than static shortcuts.

\section{Reliability Under Business Variance}
\label{app:variance}

Long-horizon business outcomes are inherently variable. Early decisions change the capital, inventory, and opportunities available later, allowing small differences between agent trajectories to compound before final scoring. For trustworthy conclusions, we evaluate whether the conclusions of Business Arena persist across repeated runs.

\subsection{Ranking Stability Under Repeated Sampling}
\label{app:variance-ranking}

Each of the 15 models is evaluated in ten separately launched episodes under the same world seed and experimental configuration. The leaderboard uses mean final net worth across these runs. To test whether its main ordering depends on a few exceptional trajectories, we conduct a split-half resampling analysis. For each replicate, we independently partition the ten runs of every model \(m\) into two disjoint sets \(A_m\) and \(B_m\), each containing five runs, and compute
\[
\bar{s}^{A}_m
=
\frac{1}{5}\sum_{r\in A_m}s_{mr},
\qquad
\bar{s}^{B}_m
=
\frac{1}{5}\sum_{r\in B_m}s_{mr}.
\]
We rank the 15 models independently using the two sets of means and calculate their Spearman rank correlation. We repeat this procedure over 50,000 random joint partitions, recording the rank correlation, agreement on the leading three-model group, and agreement on the exact leader. Because no trajectory appears in both halves, agreement indicates that persistent differences between models survive the particular runs used to estimate them.

The two five-run rankings achieve a mean Spearman correlation of \(0.898\) and recover the same leading three-model group in \(98.7\%\) of partitions. The main ranking therefore reflects persistent model strength rather than being driven by a few exceptional trajectories.

\begin{figure*}[t]
    \centering
    \captionsetup{
        font={small,normalfont},
        labelfont=bf,
        textfont=normalfont
    }
    \includegraphics[width=\textwidth]
        {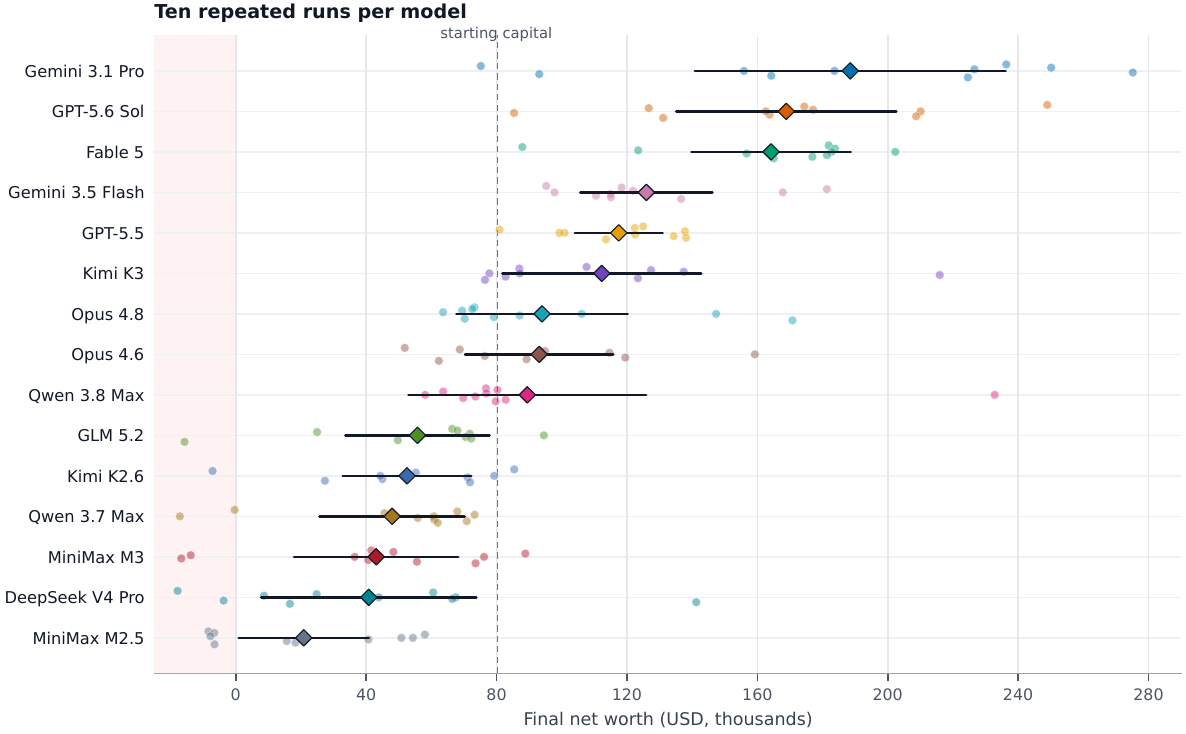}
    \caption{\textbf{Repeated-run leaderboard performance.}
    Each point represents one finalized episode; diamonds denote model means,
    and horizontal lines show two-sided \(95\%\) confidence intervals over ten
    runs. The dashed line marks the \$80{,}000 starting capital. Persistent
    differences between models remain visible despite meaningful variation
    across repeated runs of the same model.}
    \label{fig:variance-stability}
\end{figure*}

\subsection{Between- and Within-Model Variation}
\label{app:variance-components}

Figure~\ref{fig:variance-stability} shows that repeated trajectories of the same model can produce economically different outcomes, even when broader differences between models remain clear. A fixed-effects decomposition provides a descriptive summary: model identity explains \(64.8\%\) of the observed score variation, while \(35.2\%\) remains among repeated runs of the same model.

We further estimate the reliability of repeated evaluation using a one-way random-effects model. Let the final net worth of model \(m\) in run \(r\) be
\[
s_{mr}
=
\mu+\alpha_m+\epsilon_{mr},
\]
where \(\alpha_m\) represents a persistent model-level deviation and \(\epsilon_{mr}\) captures within-model trajectory variation. For the balanced design with \(R=10\) runs per model, the variance components are estimated as
\[
\widehat{\sigma}_{\mathrm{within}}^2
=
MS_{\mathrm{within}},
\qquad
\widehat{\sigma}_{\mathrm{between}}^2
=
\frac{
MS_{\mathrm{between}}-MS_{\mathrm{within}}
}{R}.
\]
The resulting reliability of a single run is
\[
\mathrm{ICC}(1,1)
=
\frac{
\widehat{\sigma}_{\mathrm{between}}^2
}{
\widehat{\sigma}_{\mathrm{between}}^2
+
\widehat{\sigma}_{\mathrm{within}}^2
}
=
0.626.
\]
Thus, a single episode contains substantial trajectory-level variation. When performance is estimated using the mean of \(k\) runs, the within-model contribution decreases by a factor of \(k\):
\[
\mathrm{ICC}(1,k)
=
\frac{
\widehat{\sigma}_{\mathrm{between}}^2
}{
\widehat{\sigma}_{\mathrm{between}}^2
+
\widehat{\sigma}_{\mathrm{within}}^2/k
}
=
\frac{
k\,\mathrm{ICC}(1,1)
}{
1+(k-1)\mathrm{ICC}(1,1)
}.
\]
This yields \(\mathrm{ICC}(1,5)=0.893\) for a five-run mean and \(\mathrm{ICC}(1,10)=0.944\) for the reported ten-run mean. Repeated evaluation therefore substantially reduces trajectory noise and provides a highly reliable estimate of model performance under the shared leaderboard world. The agreement between this result and the empirical split-half analysis further supports the stability of the main performance ranking.

Table~\ref{tab:model-run-variance} reports the uncertainty for each model. Within-model spread remains economically meaningful even when mean differences are clear. Gemini 3.1 Pro achieves the highest mean but falls below starting capital in one run, whereas GPT-5.6 Sol, Fable 5, Gemini 3.5 Flash, and GPT-5.5 preserve their initial capital in every trial. Reliability is therefore part of business capability rather than merely an error bar around average profitability.

\begin{table*}[t]
\centering
\small
\renewcommand{\arraystretch}{1.08}
\begin{tabular}{@{}lrrrr@{}}
\toprule
\textbf{Model}
& \textbf{Mean}
& \textbf{Within-model SD}
& \textbf{95\% CI}
& \textbf{Capital preserved} \\
\midrule
Gemini 3.1 Pro
& \$188{,}488 & \$66{,}641
& [\$140{,}816, \$236{,}160] & 9/10 \\
GPT-5.6 Sol
& \$168{,}867 & \$47{,}185
& [\$135{,}113, \$202{,}620] & 10/10 \\
Fable 5
& \$164{,}204 & \$34{,}141
& [\$139{,}781, \$188{,}627] & 10/10 \\
Gemini 3.5 Flash
& \$125{,}952 & \$28{,}321
& [\$105{,}692, \$146{,}212] & 10/10 \\
GPT-5.5
& \$117{,}481 & \$18{,}816
& [\$104{,}021, \$130{,}941] & 10/10 \\
Kimi K3
& \$112{,}278 & \$42{,}644
& [\$81{,}773, \$142{,}784] & 8/10 \\
Opus 4.8
& \$93{,}946 & \$36{,}729
& [\$67{,}672, \$120{,}220] & 4/10 \\
Opus 4.6
& \$93{,}066 & \$31{,}759
& [\$70{,}347, \$115{,}785] & 6/10 \\
Qwen 3.8 Max
& \$89{,}423 & \$50{,}974
& [\$52{,}958, \$125{,}887] & 3/10 \\
GLM 5.2
& \$55{,}742 & \$30{,}921
& [\$33{,}623, \$77{,}862] & 1/10 \\
Kimi K2.6
& \$52{,}533 & \$27{,}568
& [\$32{,}812, \$72{,}254] & 1/10 \\
Qwen 3.7 Max
& \$47{,}956 & \$31{,}145
& [\$25{,}676, \$70{,}236] & 0/10 \\
MiniMax M3
& \$43{,}064 & \$35{,}218
& [\$17{,}871, \$68{,}258] & 1/10 \\
DeepSeek V4 Pro
& \$40{,}804 & \$46{,}130
& [\$7{,}805, \$73{,}803] & 1/10 \\
MiniMax M2.5
& \$20{,}856 & \$27{,}897
& [\$900, \$40{,}813] & 0/10 \\
\bottomrule
\end{tabular}
\caption{\textbf{Repeated-run variation across the 150-run cohort.}
Confidence intervals use two-sided \(t\)-intervals over ten runs. Capital
preservation counts runs ending at or above the \$80{,}000 starting capital.}
\label{tab:model-run-variance}
\end{table*}

These analyses evaluate reliability conditional on the shared world used by the leaderboard. Robustness across different market realizations is a separate question that requires crossing models with multiple world seeds. Under the present evaluation setting, the results establish that the reported performance bands primarily reflect persistent differences between models rather than isolated successful trajectories.

\section{Seller-Buyer Interface}

\begin{figure*}[h]
    \centering
    \includegraphics[width=\textwidth]{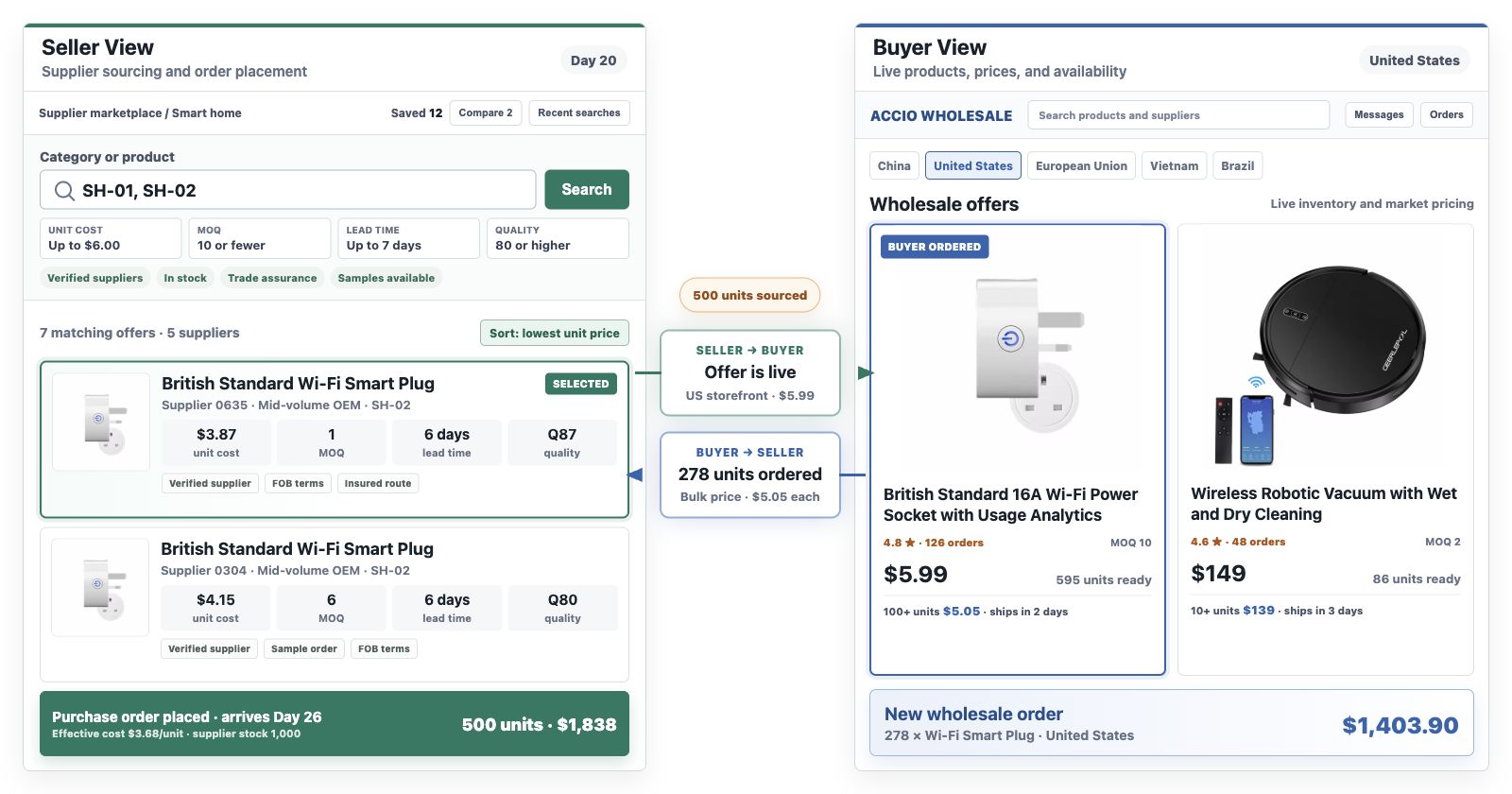}
    \caption{\textbf{Seller and buyer views in Business Arena.}
    An agent sources and lists a product while buyers discover and purchase the resulting offer.}
    \label{fig:market-exchange}
\end{figure*}

Figure \ref{fig:market-exchange} gives an example of the seller-buyer interface in Business Arena (note that this is a dedicated illustration of the marketplace, business arena involve no evaluation of GUI agent use). 

\section{Expert-designed Strategies}
\label{app:strategy-reserve}

\paragraph{Design objective.}
The strategy reserve estimates what a competent operator can achieve from the same evidence available to evaluated agents. Its policies interact with the arena exclusively through the public interface and cannot access hidden demand parameters, future events, simulator seeds, or database internals. More importantly, the reserve is not constructed by independently selecting the best-performing rule for each arena feature. We represent business operation as a hierarchy of coupled decisions in which upstream beliefs and plans constrain downstream actions, and realized outcomes revise the next operating cycle.

\begin{figure*}[t]
    \centering
    \includegraphics[width=0.94\linewidth]{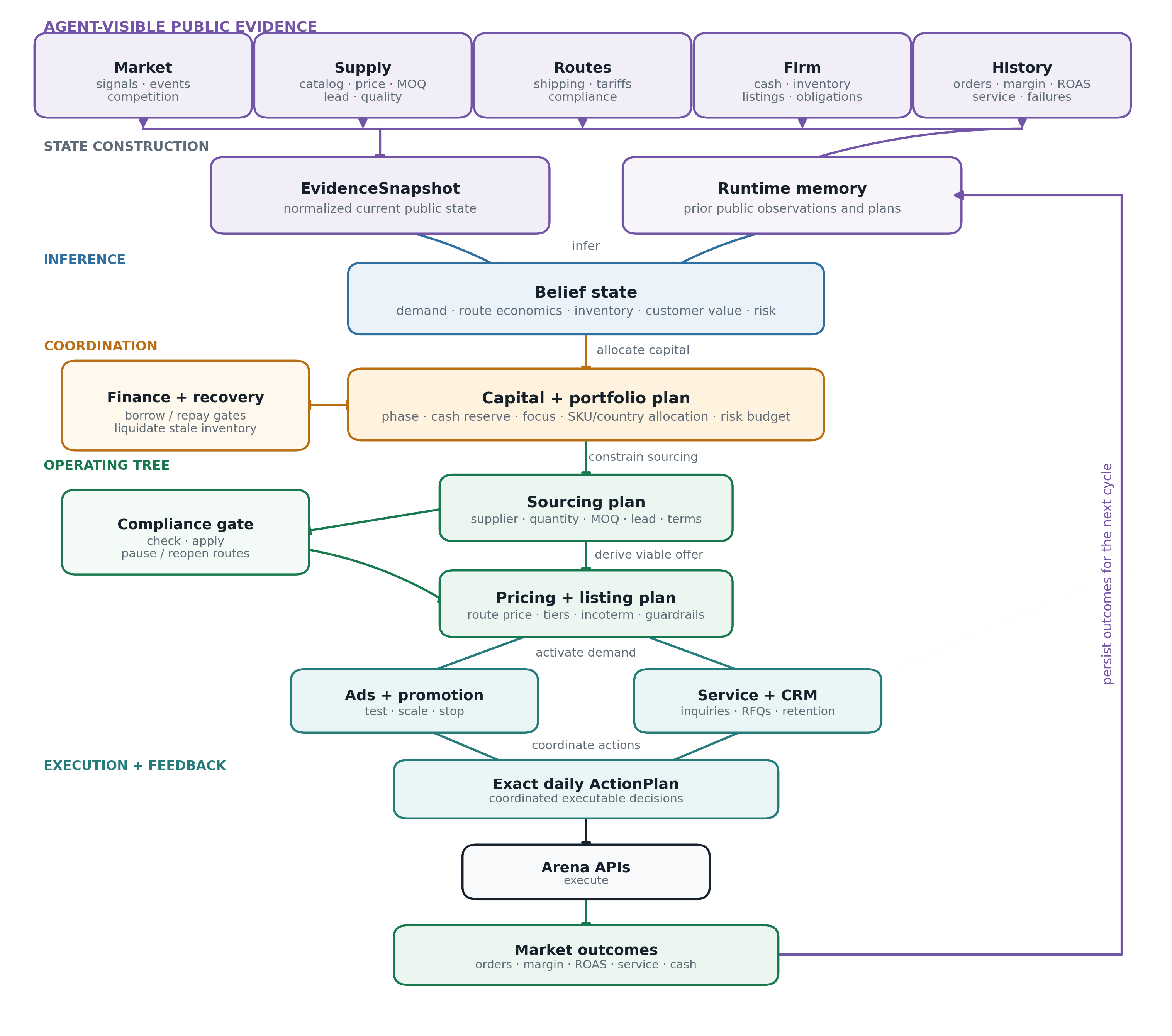}
    \caption{\textbf{Expert-designed strategy decision architecture.}
    Agent-visible evidence and prior operating history are converted into beliefs of the market, which inform a coordinating capital and portfolio plan. This plan constrains the downstream operating tree, whose actions are executed through the public arena interface. Realized outcomes are retained as public memory and revise both beliefs and resource allocation in the next cycle.}
    \label{fig:strategy-reserve-pipeline}
\end{figure*}

\paragraph{Public evidence and memory.}
At the beginning of each cycle, the controller constructs a normalized evidence snapshot from five public information groups: the external market, available supply, route economics, the firm's current state, and realized operating history. A bounded runtime memory retains earlier observations and plans, including market anchors, sales and inventory histories, advertising outcomes, customer interactions, and prior portfolio decisions. This memory contains only information previously returned through the public interface; it does not reconstruct unobserved state from the simulator.

\paragraph{Belief construction.}
Current evidence and memory are converted into an explicit belief state. Demand beliefs combine public market signals with realized velocity and sell-through; route beliefs compose supplier cost, shipping, tariffs, competition, and other visible costs; inventory beliefs distinguish productive positions from stale or weak stock; and customer-value beliefs summarize observed service and account outcomes. The controller also records uncertainty, allowing early decisions to remain exploratory when little realized evidence is available.

\paragraph{Capital and portfolio coordination.}
The capital and portfolio plan is the coordinating layer of the reserve. It determines the operating phase, cash reserve, deployable budget, portfolio breadth, product and country allocation, and acceptable risk exposure. Its behavior changes with feedback: early cycles fund diversified probes, demonstrated winners become eligible for scaling, weak portfolios trigger reduced deployment and repair, and late cycles prioritize cash recovery. Consequently, specialized modules cannot spend independently; sourcing, advertising, financing, and recovery decisions must remain consistent with a shared resource plan.

\paragraph{Operating tree.}
The portfolio plan first constrains sourcing by specifying which opportunities may receive capital and at what depth. Sourcing then determines the feasible cost, quality, lead time, and inventory position from which route-specific prices and listings are derived. Compliance acts as a gate on planned routes: required approvals must be identified and initiated before affected offers can operate. Advertising and promotion are activated only after inventory and offer economics are viable, while customer operations respond to demand reaching the resulting offer. Finance and liquidation protect the shared capital plan by supporting justified expansion or recovering cash from stale positions.

\begin{table*}[t]
\centering
\small
\begin{tabular}{p{0.15\textwidth} p{0.34\textwidth} p{0.43\textwidth}}
\toprule
\textbf{Component} & \textbf{Strategy coverage} & \textbf{Connection to the decision system} \\
\midrule
Evidence and memory
& Market signals, events, competition, supplier offers, route costs, firm state, and realized operating history
& Normalizes current public evidence and retains bounded observations and prior plans for the next cycle. \\
\addlinespace

Opportunity beliefs
& Market-depth, trend- and event-aware, realized-velocity, unit-economics, inventory-risk, and customer-value signals
& Converts heterogeneous evidence into demand, route, inventory, and customer beliefs consumed by the portfolio planner. \\
\addlinespace

Capital and portfolio
& Diversified probing, conviction-weighted deployment, broad velocity portfolios, compact capital-efficient books, adaptive focus, and recovery phases
& Sets the cash reserve, risk budget, portfolio scope, and per-opportunity allocation that constrain all downstream spending. \\
\addlinespace

Sourcing
& Cost-balanced, fast-turn, quality-led, risk-adjusted, and relationship-aware supplier selection
& Selects supplier, quantity, timing, and terms within the portfolio allocation; fulfillment and realized quality revise supplier eligibility. \\
\addlinespace

Pricing and listing
& Margin-preserving, competition-aware, volume-oriented, and premium offer policies with route-specific cost floors
& Translates sourced inventory and route economics into viable offers; conversion, margin, and inventory age update future prices. \\
\addlinespace

Compliance
& Route checks, permit application, temporary listing pauses, and reopening after approval
& Gates sourcing and listing plans before trade; permit status, violations, and fines feed operational reliability. \\
\addlinespace

Ads and promotion
& Bounded experimentation, test--scale--stop rules, event-timed promotion, and recovery shutdown
& Operates only on viable, inventory-backed offers; full-funnel returns affect advertising and replenishment decisions. \\
\addlinespace

Customer operations
& Inquiry and RFQ handling, fulfillment-aware responses, conservative negotiation, and, where enabled, retention-oriented CRM
& Converts incoming demand while protecting feasibility and contribution; service outcomes update customer-value and demand beliefs. \\
\addlinespace

Finance and recovery
& Cash reserves, evidence-gated borrowing and repayment, position limits, and stale-inventory liquidation
& Expands deployment only when supported by visible economics and returns capital when continued ownership is no longer justified. \\
\addlinespace

Feedback control
& Sell-through learning, winner scaling, loser pauses, route adaptation, focus revision, and portfolio rebalancing
& Routes realized orders, margins, stock, advertising, service, and failures back to both beliefs and capital allocation. \\
\bottomrule
\end{tabular}
\caption{\textbf{Coverage and coupling of strategy-reserve components.}
Each component declares the public evidence or upstream plan it consumes, the decision object it produces, and the realized feedback that can revise it.}
\label{tab:strategy-reserve-components}
\end{table*}

\paragraph{Successful human-designed strategies.}
In this section we show a few examples of what human-designed strategies could succeed in practice.

\textbf{Bayesian export compounder.} This strategy treats each SKU--country route as an uncertain opportunity and updates its beliefs using public market observations and realized outcomes. It maintains a diversified CN--EU--US portfolio, shifts shared inventory toward routes with stronger evidence of contribution, and coordinates pricing, replenishment, advertising, customer relationships, supplier terms, and credit so that profitable sales fund subsequent expansion.

\textbf{Relationship-led wholesaler.} This strategy builds value through durable supplier and buyer relationships. It concentrates capital on products supported by repeat demand, negotiates better sourcing terms, protects landed margin, and redeploy capitol only when realized turnover can support them.

\textbf{Velocity market-maker.} This strategy prioritizes liquidity and inventory turnover. It maintains broad product coverage, favors fast replenishment, uses volume-oriented pricing, and scales only products and advertisements that convert inventory into cash quickly.

\textbf{Event swing trader.} This strategy reacts to time-sensitive opportunities. It combines public event signals with observed sales, lead times, and remaining event duration, then coordinates sourcing, pricing, focus, and promotion around opportunities that are confirmed by market response.

\textbf{Inventory-light broker.} This strategy minimizes ownership risk. It opens small positions in products with favorable contribution per unit of committed capital, favors short routes and low minimum orders, avoids paid acquisition, and replenishes only after sales evidence appears.

\textbf{Premium account house.} This strategy competes through quality and account value. It selects stronger suppliers, maintains a focused assortment, charges margin-protective prices, and uses reviews, refunds, and repeat-buyer behavior to decide whether to invest further or exit.

\begin{figure*}[t]
    \centering
    \includegraphics[width=\textwidth]{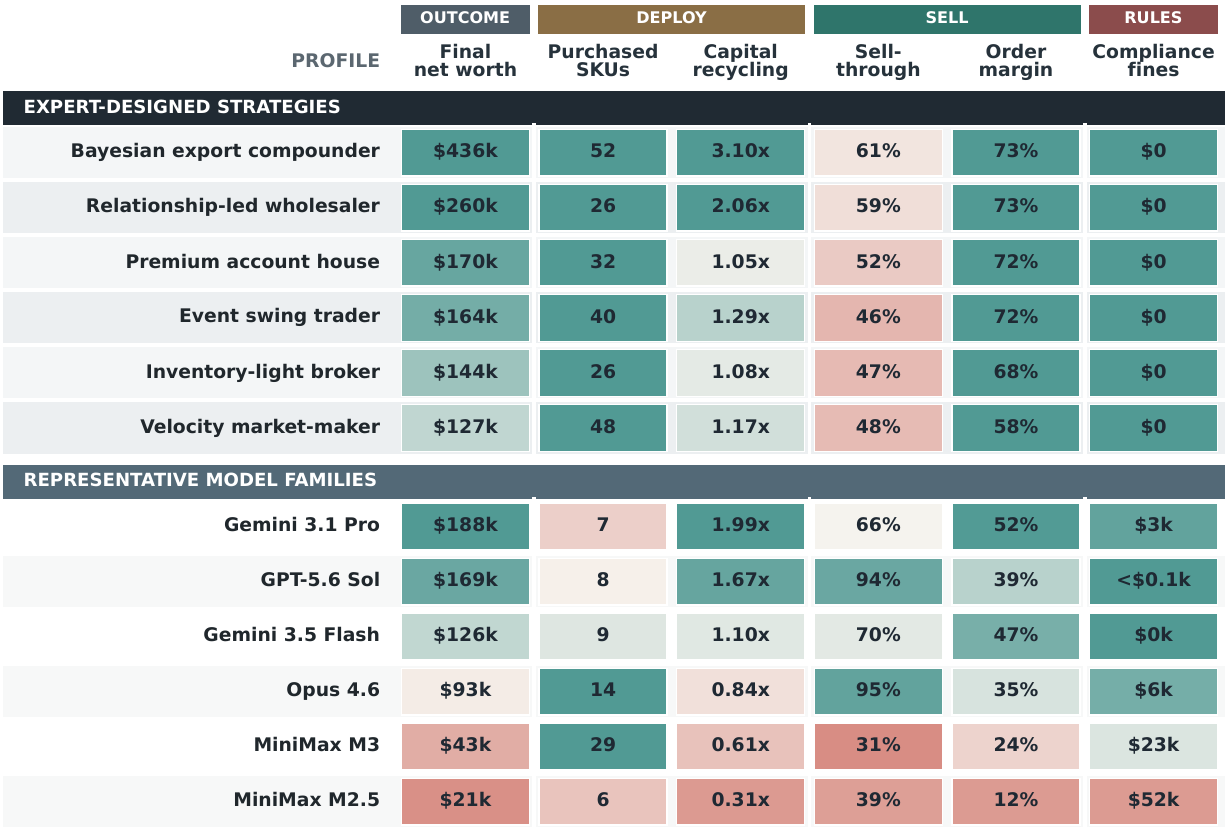}
    \caption{ Comparison of expert-designed strategies and selected models. The strongest expert strategies recycle returned cash across more SKUs while preserving higher margins and avoiding compliance fines.}
    \label{fig:strategy-model-operating-profiles}
\end{figure*}

\paragraph{Why expert-designed strategies win.}
Figure~\ref{fig:strategy-model-operating-profiles} shows the skill-level comparison between expert-designed strategies and models. The strongest strategies repeatedly reinvest returned cash, recycling capital by $2.06$--$3.10$ times, compared with $1.67$--$1.99$ times for the two leading model families; this compounding routine is central to wholesale practice because early sales finance later inventory rather than leaving the business constrained by its initial budget. They also purchase across 26--52 SKUs, while the leading models use only 7--8, suggesting that the arena rewards the realistic work of searching broadly for opportunities and diversifying demand risk. The comparison further reveals a margin--sell-through trade-off: expert strategies accept 46--61\% sell-through while preserving 58--73\% order margins, whereas GPT-5.6 Sol and Opus 4.6 reach 94--95\% sell-through but only 35--39\% margins; despite moving inventory quickly, both remain below the two highest-performing expert strategies in final net worth. Rapid liquidation is therefore not sufficient when prices satisfy too much contribution per order. Finally, every expert strategy incurs zero compliance fines, showing that strong operation treats routine compliance checks as hard gates on otherwise profitable actions.

\section{Stateful Evaluation and Controlled Continuations}
\label{app:stateful-evaluation}

\subsection{Implementation}

Borrowing the state taxonomy from checkpoint/restore literature~\cite{ansel2009dmtcp,dong2026deltabox,wu2026crab}, Business Arena requires filesystem and application-semantic restoration rather than exact process-level restoration. Its services can restart, but each branch must recover the marketplace and workspace together with the exact model-visible context and tool boundary for the next request. The marketplace fits in a transactional database checkpoint; the harder harness state cannot be recovered exactly from trajectory logs and workspace files alone.

We therefore use an in-house white-box harness that checkpoints the exact post-compaction message sequence, tool-operation boundary, harness and model provenance, and a content-addressed workspace archive. A child verifies all digests, restores the marketplace and workspace, and reconstructs the continuation before tool discovery or its first model request, with the system message installed exactly once and checkpoint internals hidden from the model.

\subsection{Trace Search Case Study}

\begin{figure*}[t]
    \centering
    \includegraphics[width=\linewidth]{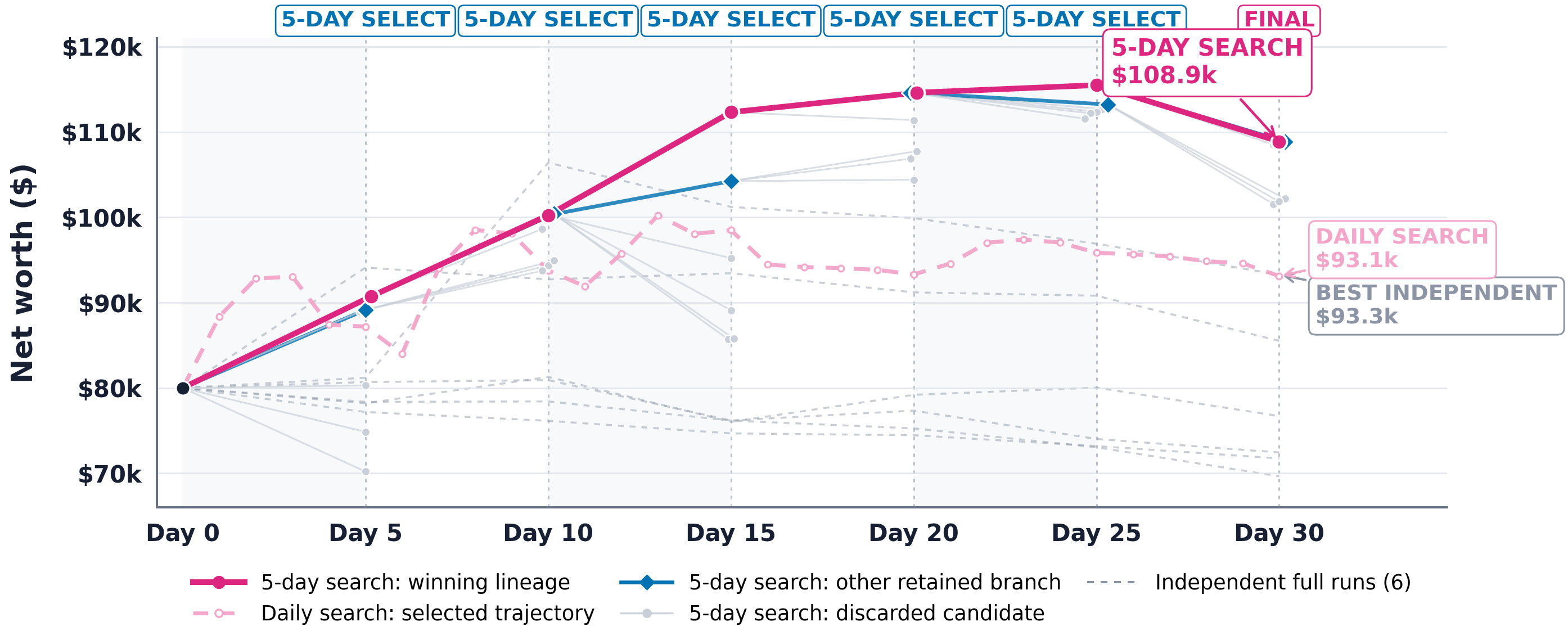}
    \caption{\textbf{Checkpoint cadence matters when scaling long-horizon business operation.} The left panel shows the continuations considered by five-day trace search, its retained lineages, the trajectory retained by daily search, and six independent full runs. Intermediate values through Day 25 are checkpoint-evaluator estimates; Day 30 reports exact final net worth after liquidation. All conditions use the same model, world, prompt, and 180-model-day rollout budget.}
    \label{fig:trace-search-cadence}
\end{figure*}

Stateful evaluation makes it possible to allocate more test-time compute for better performance. We study this through \emph{trace search}: every retained trace is forked into \(M\) continuations, each continuation runs for \(H\) arena days, and the global top \(K\) states are retained for the next segment. Figure~\ref{fig:trace-search-cadence} reports results on Qwen 3.8 Max Preview with \(K=2\), \(M=3\). Five-day forking reaches a final net worth of \$108{,}878; forking and selecting every day reaches \$93{,}111; and the best of six independent 30-day runs (with the same budget) reaches \$93{,}268. Thus, five-day trace search finishes \(16.9\%\) above daily trace search and \(16.7\%\) above the strongest independent run under the same model-day budget.

The independent-run comparison shows how resumability can turn intermediate business states into useful test-time-compute allocation: in this case, branching from retained superior states is more effective than spending the same rollout budget on independent episodes. More importantly, the comparison provides direct evidence that the arena is long-horizon with delayed feedback. The branch that appears strongest after one day does not lead to the strongest final business, because sourcing, delivery, demand response, repricing, and inventory turnover reveal their consequences over longer intervals. Daily selection therefore fails to compose locally attractive states into the best global trajectory, while five-day selection preserves enough horizon for delayed outcomes to become informative.

\section{Model Diagnostic Profiles}
\label{app:model-diagnostics}

Final net worth identifies overall performance but does not reveal where a model succeeds or fails. Figure~\ref{fig:model-diagnostic-heatmap} summarizes operational strengths and weaknesses across the business cycle. Each cell reports the model's mean over repeated runs, with color indicating its cohort-relative standing. We provide below a detailed explanation of these metrics.

\subsection{Operating Fluency}

\paragraph{Arena calls.}
We count all arena operations, including those issued through model-authored scripts. More calls generally indicate greater fluency and engagement with the environment, while few calls suggest limited operation. 

\paragraph{World checks.}
World-check rate measures how frequently the agent refreshes changing information such as market conditions, events, and tariffs. A high rate indicates active monitoring, whereas a low rate suggests that decisions may rely on stale assumptions.

\paragraph{Call failures.}
Call-failure rate is the fraction of arena operations that return unsuccessful results. A low rate reflects reliable interface use; a high rate reveals malformed calls, infeasible actions, or repeated execution mistakes.

\subsection{Capital Deployment}

\paragraph{Capital utilization.}
Capital utilization measures the peak value deployed into inventory, receivables, and escrow relative to starting capital. High utilization indicates active deployment and may exceed \(100\%\) when capital is recycled. Low utilization indicates idle capital, although excessive deployment can still produce stranded inventory.

\paragraph{Holding cost.}
Holding cost measures inventory-carrying fees relative to sourcing expenditure. A low ratio indicates disciplined inventory turnover, while a high ratio suggests oversized or slow-moving positions.

\subsection{Selling Performance}

\paragraph{Sell-through.}
Sell-through is the share of available inventory sold by the end of the episode. High sell-through indicates effective demand capture; low sell-through reveals weak sales or stranded inventory.

\paragraph{Order margin.}
Order margin measures the share of revenue remaining after product cost, shipping, tariffs, and insurance. A high margin reflects pricing that accounts for the full cost stack, while a low margin indicates underpricing or omitted costs.

\paragraph{Route cost.}
Route cost measures seller-paid shipping and tariffs relative to revenue. A low ratio indicates efficient route selection or successful cost allocation, whereas a high ratio means cross-border costs consume substantial value.

\paragraph{Advertising return.}
Advertising return compares contribution attributed to ads against advertising spending. Positive values indicate profitable demand acquisition; low or negative values indicate that advertising fails to recover its cost.

\paragraph{Market share.}
Market share is the fraction of marketplace orders captured by the agent. A high share indicates strong demand capture relative to NPC competitors, while a low share suggests that competing sellers win most available demand.

\subsection{Customer Interaction}

\paragraph{Buyer conversion.}
Buyer conversion is the share of answered inquiries that result in purchases. A high rate indicates effective inference of customer preferences and viable offer construction; a low rate means that replies rarely close sales.

\paragraph{Factual replies.}
Factual-reply quality measures how often model responses avoid materially false claims. Strong models provide complete, evidence-grounded product and fulfillment information, while weak models make unsupported or incorrect claims.

\paragraph{RFQ success.}
RFQ success is the share of received RFQs that produce accepted orders. A high rate reflects effective negotiation of feasible bulk offers; a low rate indicates missed, rejected, or operationally infeasible opportunities.

\subsection{Compliance Reliability}

\paragraph{Fines.}F
Fines sum the monetary penalties caused by compliance violations. Low or zero fines indicate that the agent turns visible requirements into reliable operating gates. High fines reveal unsafe operation that can overwhelm otherwise plausible commercial performance.

\section{Model Behavior Comparisons}
\label{app:model_comparisons}

In this section we provide a more refined analysis of model behaviors, containing numerical gains/losses traceable back to each action chains.

\begin{figure*}[h]
    \centering
    \includegraphics[width=\linewidth]{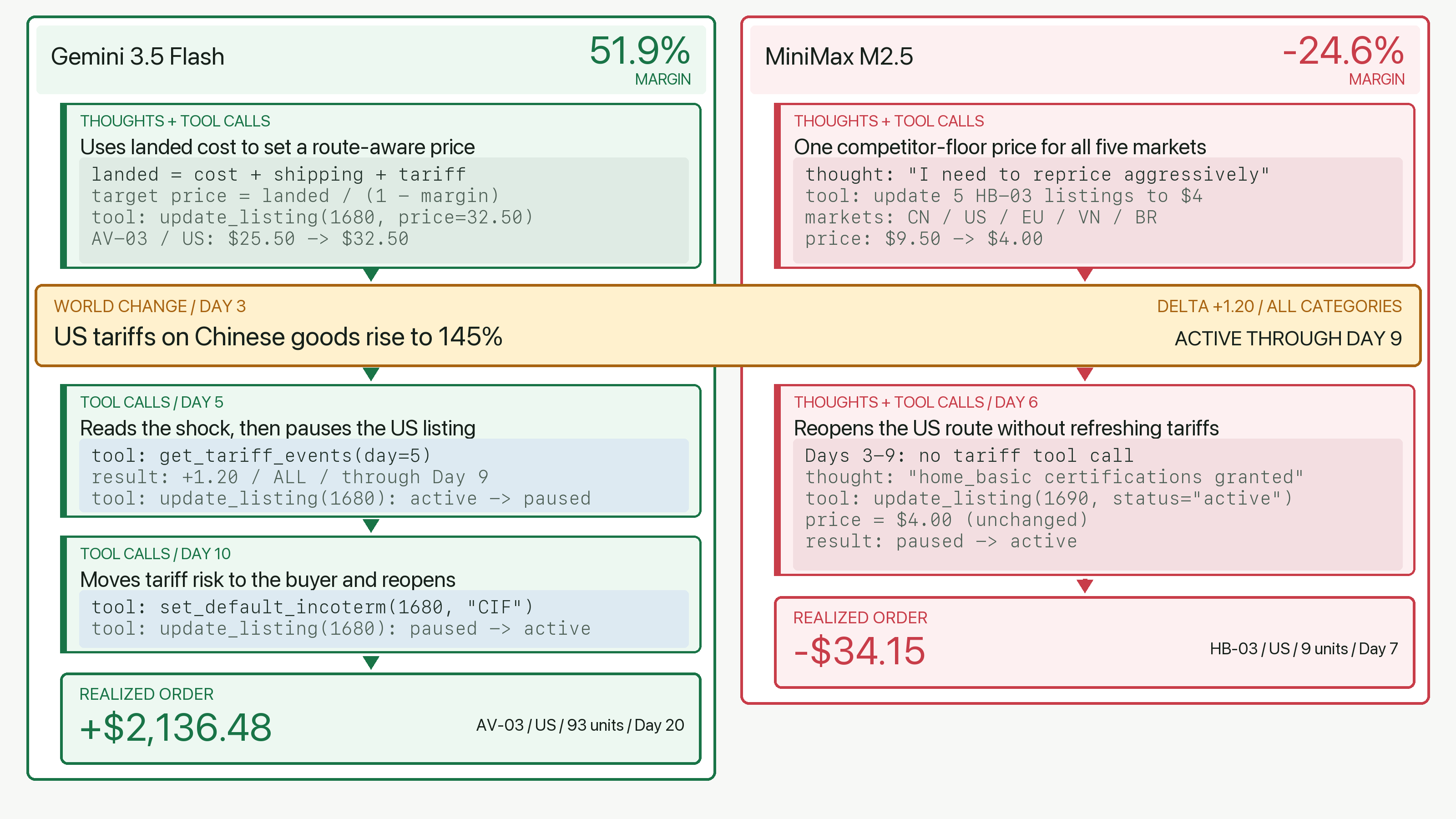}
    \caption{\textbf{Different decisions under the same market shock.} Gemini 3.5 Flash incorporates the tariff change into route-specific pricing and market access, while MiniMax M2.5 reopens the affected route without refreshing its cost assumptions. Each decision is linked to its realized order outcome.}
    \label{fig:pricing-action-trace}
\end{figure*}

\begin{figure}
    \centering
    \includegraphics[width=\linewidth]{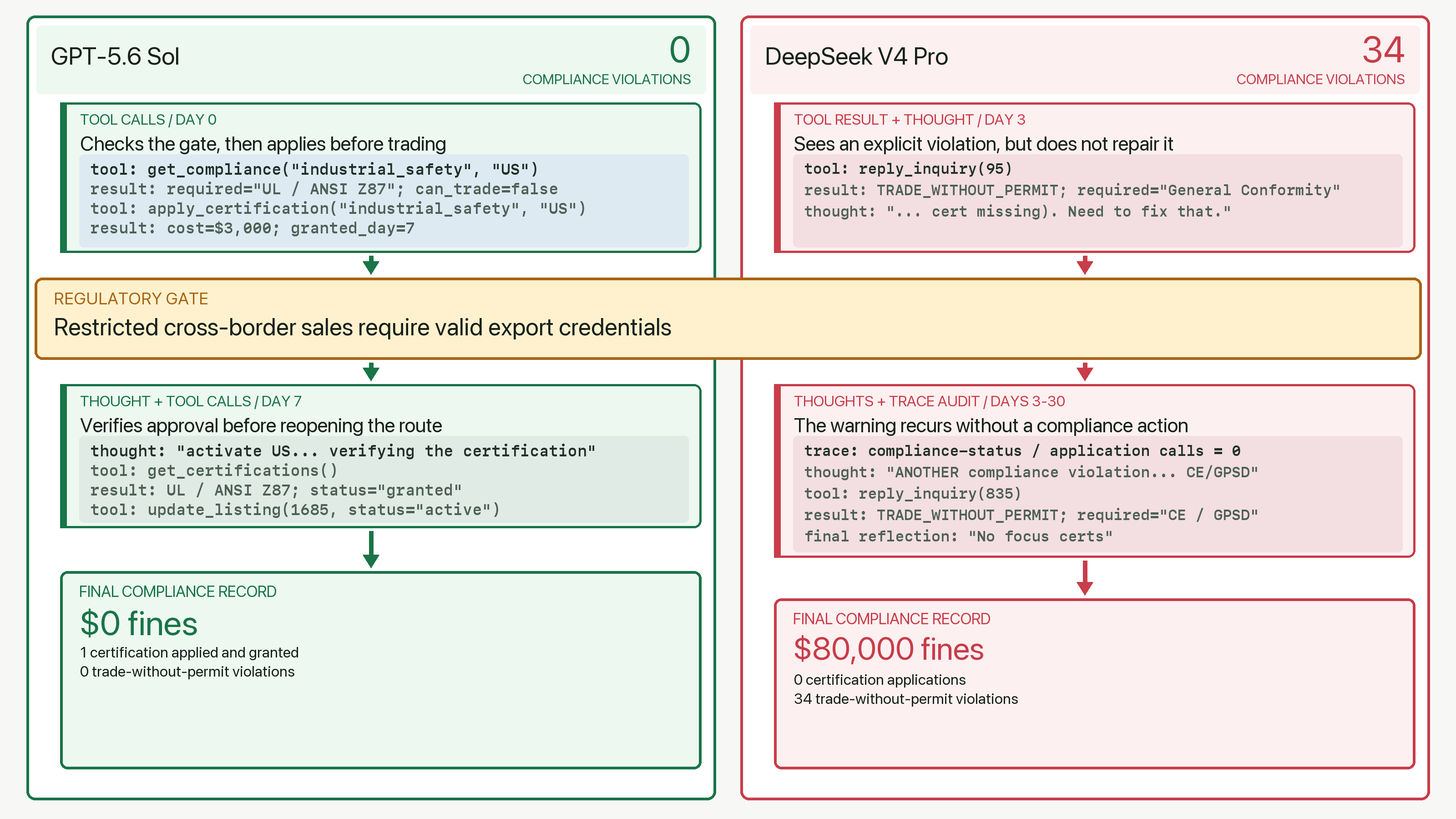}
    \caption{\textbf{Compliance behavior of GPT-5.6 Sol and DeepSeek V4 Pro.} GPT-5.6 Sol checks certification requirements, applies for the required credential, and verifies approval before reopening trade. DeepSeek V4 Pro recognizes repeated compliance violations but does not take corrective action, resulting in substantial fines.}
    \label{fig:compliance-action-trace}
\end{figure}

\paragraph{Compliance.}
Figure~\ref{fig:compliance-action-trace} contrasts how two models close the loop between regulatory evidence and action. GPT-5.6 Sol checks the applicable certification requirement, applies for the missing UL/ANSI credential, and verifies that it has been granted before reopening the US listings, completing the episode with zero compliance violations and no fines. DeepSeek V4 Pro receives an explicit trade-without-permit warning and recognizes in its reasoning that the missing certification must be fixed, but never calls the compliance-check or certification-application tools. The same failure later recurs for other routes, resulting in 34 violations and \$80{,}000 in fines. The comparison distinguishes merely recognizing a compliance problem from reliably converting that recognition into corrective action.

\begin{figure}
    \centering
    \includegraphics[width=\linewidth]{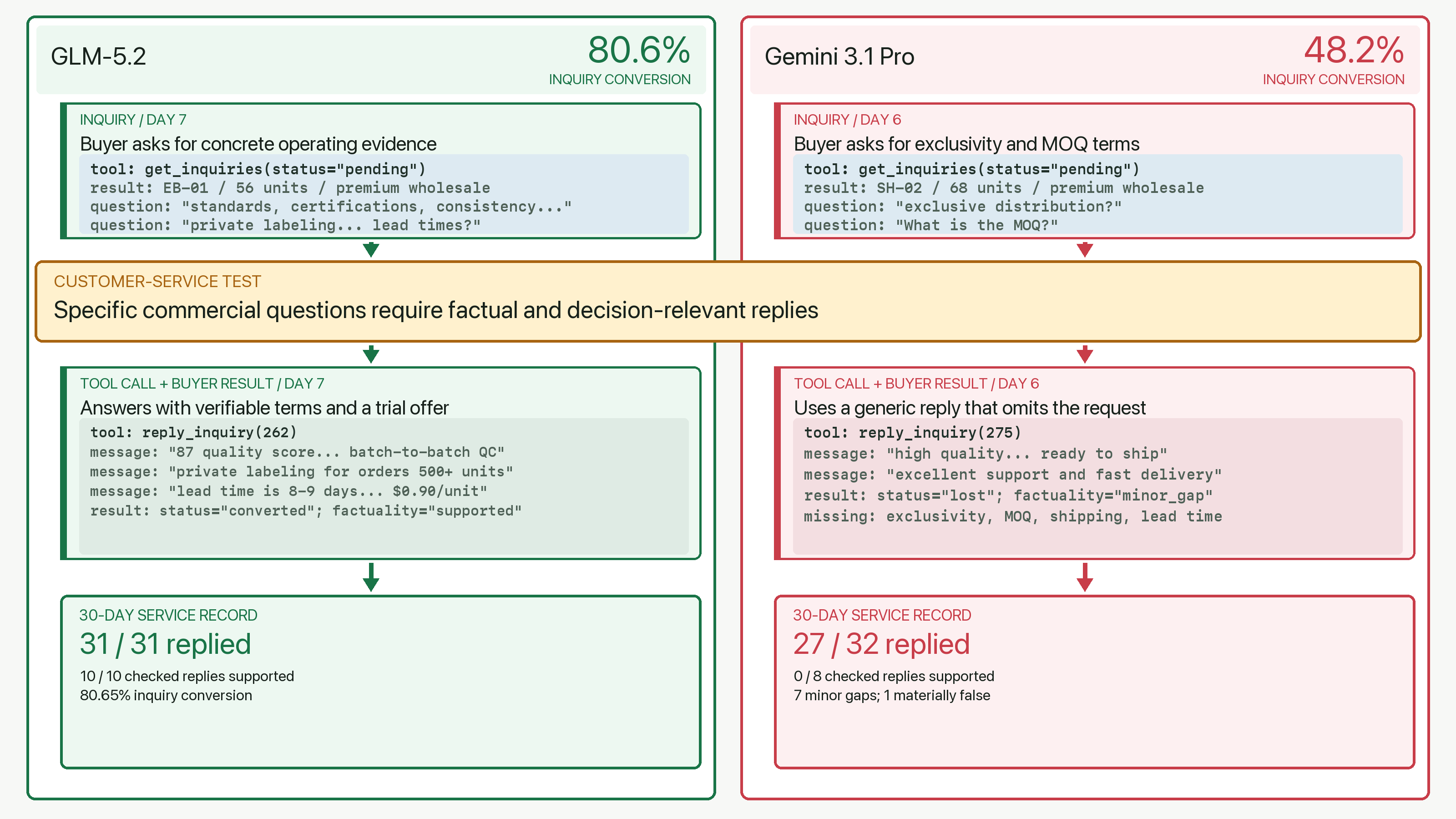}
    \caption{\textbf{Customer-service behavior of GLM-5.2 and Gemini 3.1 Pro.} GLM-5.2 addresses the buyer's specific commercial questions with concrete, factually-supported information, while Gemini 3.1 Pro gives a generic response that omits key decision criteria and loses the inquiry.}
    \label{fig:customer-service-action-trace}
\end{figure}

\paragraph{Customer service.}
Figure~\ref{fig:customer-service-action-trace} illustrates the importance of factual and decision-relevant communication. When asked about fact-related questions in the customer service subtask, GLM-5.2 provides concrete product and commercial terms; the response is evaluated as supported and converts the inquiry. Across the episode, it replies to all 31 inquiries and achieves an 80.65\% conversion rate. Gemini 3.1 Pro instead answers requests using largely templated responses, leaving the buyer's central questions unanswered and losing the inquiry. It replies to 27 of 32 inquiries, achieves 48.15\% conversion, and has no fully supported response among the eight checked replies. This example shows the design philosophy of the customer-service task: response count/timeliness alone does not measure customer-service ability; specificity, factual support, and coverage of the buyer's decision criteria matters more in reality.

\begin{figure}
    \centering
    \includegraphics[width=\linewidth]{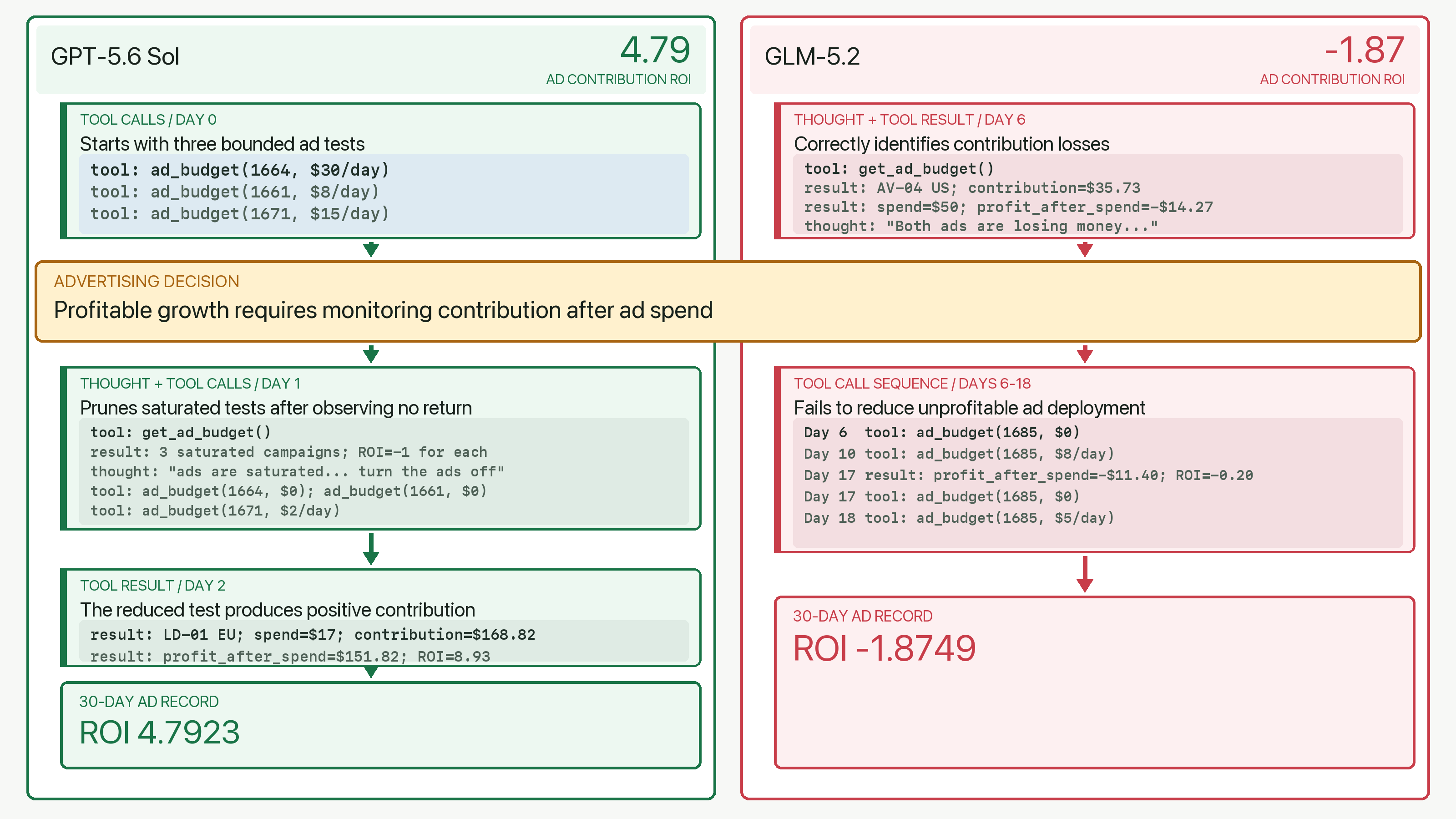}
    \caption{\textbf{Advertising strategies of GPT-5.6 Sol and GLM-5.2.} Both models observe contribution-level feedback, but GPT-5.6 Sol adapts its advertising allocation by pruning unprofitable campaigns and refining its strategies. GLM-5.2 recognizes negative returns but does not consistently translate this feedback into sustained budget adjustments.}
    \label{fig:ads-action-trace}
\end{figure}

\paragraph{Advertising.}
Figure~\ref{fig:ads-action-trace} compares whether observed advertising returns are translated into disciplined budget allocation. GPT-5.6 Sol begins with three bounded tests, observes that all three initially have negative contribution ROI, stops two campaigns, and reduces the remaining campaign to \$2 per day. The reduced test subsequently produces \$151.82 in profit after ad spend and an ROI of 8.93, while the episode ends with an aggregate advertising ROI of 4.79. GLM-5.2 also identifies that an advertised route is unprofitable, but repeatedly stops and reactivates spending on that route without sustained improvement, ending with an ROI of -1.87 . The contrast shows the what we expect from models: trying out ads spending in small amounts initially and adapting to the market feedback, which is exactly what advertising needs in reality.

\end{document}